\documentclass[10pt,twocolumn,letterpaper]{article}

\usepackage[pagenumbers]{cvpr}      
\makeatletter
\renewcommand{\paragraph}{%
  \@startsection{paragraph}{4}{\z@}%
  {1.0ex}%
  {-1em}%
  {\normalfont\normalsize\bfseries}%
}
\makeatother

\usepackage{graphicx}
\usepackage{url}
\usepackage{hyperref}
\usepackage[numbers, sort&compress]{natbib}
\usepackage{fancyhdr}
\usepackage[utf8]{inputenc}
\usepackage[T1]{fontenc}
\usepackage{booktabs}
\usepackage{amssymb}
\usepackage{amsmath}
\usepackage{colortbl}
\usepackage{tabularx}
\usepackage{array}
\usepackage{geometry}
\usepackage{xspace}
\usepackage{enumitem}
\usepackage{float}

\usepackage{multirow}
\usepackage{makecell}
\usepackage[table]{xcolor}
\usepackage{threeparttable}

\newcolumntype{L}{>{\raggedright\arraybackslash}X}
\newcolumntype{M}{>{\raggedright\arraybackslash}p{3.5cm}}

\usepackage{hyperref}
\definecolor{darkblue}{rgb}{0, 0, 0.5}
\definecolor{skyblue}{rgb}{0.289, 0.484, 0.72}
\hypersetup{colorlinks=true, citecolor=skyblue, linkcolor=red, urlcolor=darkblue, bookmarksopen=false, bookmarksnumbered=true}

\definecolor{groupgray}{RGB}{241,241,241}
\definecolor{oursred}{RGB}{255,232,232}

\definecolor{mygreen}{RGB}{0,160,0}   
\definecolor{myred}{RGB}{200,0,0}

\usepackage{graphicx}
\newcolumntype{S}[1]{>{\centering\arraybackslash}m{#1}} 
\newcolumntype{L}{>{\raggedright\arraybackslash}X}
\newcolumntype{C}{>{\centering\arraybackslash}X}

\newcolumntype{P}[1]{%
  >{\raggedright\arraybackslash}p{#1}}
\newcolumntype{Y}{%
  >{\raggedright\arraybackslash}X}
  
\newcommand{\ours}{Swift-Image\xspace}

\title{Exploring the Performance Frontier of Compact Unified Image Generation Models}
\author{
\fontsize{12pt}{11pt}\selectfont
Taihang Hu$^{*}$, Zhao Wang$^{*}$, Zuan Gao$^{*}$, Tao Liu$^{*}$, Hao Yan$^{*}$, 
Zhengze Xu$^{*}$, Yuhang Yu$^{*}$, Yongchao Du$^{*}$, \\
\fontsize{12pt}{11pt}\selectfont
Xingjian Wang$^{*}$, 
Jun Zheng$^{*}$, Qinye Zhou$^{*}$, Yaqi Cai, Zhengrui Chen, Chao Lin, Yefeng Shen, Yuan Wang, \\
\fontsize{12pt}{11pt}\selectfont
Zhengtao Wu, Ge Wu, Xiaoli Xu, Denghui Yang, Huayu Zhang, Mingzhou Zhang, 
Mengting Chen$^{\dagger}$ \\
{\small $^{*}$Equal Contribution \qquad $^{\dagger}$Corresponding Author}
\vspace{1.5mm}
\\
}
\date{\today}

\begin{document}
\maketitle
\begin{abstract}
\vspace{-4mm}
We present \textbf{\ours}, a compact unified model for text-to-image generation, single-image editing, and multi-image editing. Our goal is to explore how far a relatively small visual generator can be pushed through systematic training engineering under a constrained computational budget. \ours adopts an efficient 6B single-stream DiT and a progressive training pipeline that evolves from broad semantic coverage to higher resolution, stronger visual quality, and unified generation-editing supervision. For post-training, we employ parallel expert reinforcement learning followed by multi-teacher on-policy distillation to alleviate interference among heterogeneous objectives. We further decouple high-level reasoning from pixel-level rendering with a Prompt Enhancer that translates user requests into generator-aligned visual specifications. For efficient deployment, structural pruning and few-step distillation produce \textbf{3B} and accelerated variants. \ours achieves leading aggregate performance among evaluated open-source models with \textbf{only 6B parameters and 243K GPU training hours}; the compressed 3B model incurs nearly no loss, while few-step distillation further improves aggregate editing performance with substantially fewer sampling steps. Our study also summarizes practical lessons for architecture, data curriculum, post-training, prompt enhancement, and model compression.

\vspace{-4mm}
\end{abstract}
\section{Introduction}
Recent image generation models have achieved substantial progress in visual fidelity, text rendering, instruction following, and reference-based editing \citep{esser2024scaling, wu2025qwen, blackforest2025flux2klein, cao2025hunyuanimage, gptimage2_model_card, google2025nanobanana, hu2026orion}. However, these advances are often accompanied by increasingly large generative backbones\citep{cao2025hunyuanimage, wu2025qwen}, expensive training pipelines, and separate model variants for generation and editing \citep{krea-2-2026, firered2026rededit, song2026joyai}. This raises a practical question:

\textit{How far can the performance of a compact image model be pushed through systematic training engineering under a constrained computational budget?}

This problem is particularly challenging because a unified model must support text-to-image generation, single-image editing, and multi-image editing within one set of weights. These tasks impose different requirements: text-to-image generation demands broad visual--semantic alignment, while editing additionally requires precise use of visual references and preservation of content unrelated to the requested change. Complex user requests may further involve text rendering, compositional reasoning, domain knowledge, and layout planning. Compact models are especially sensitive to these competing demands, as inefficient architectural choices, task imbalance, or unstable training can consume their limited capacity and computation.
\begin{figure}[t]
\centering
  \includegraphics[width=0.5\textwidth]{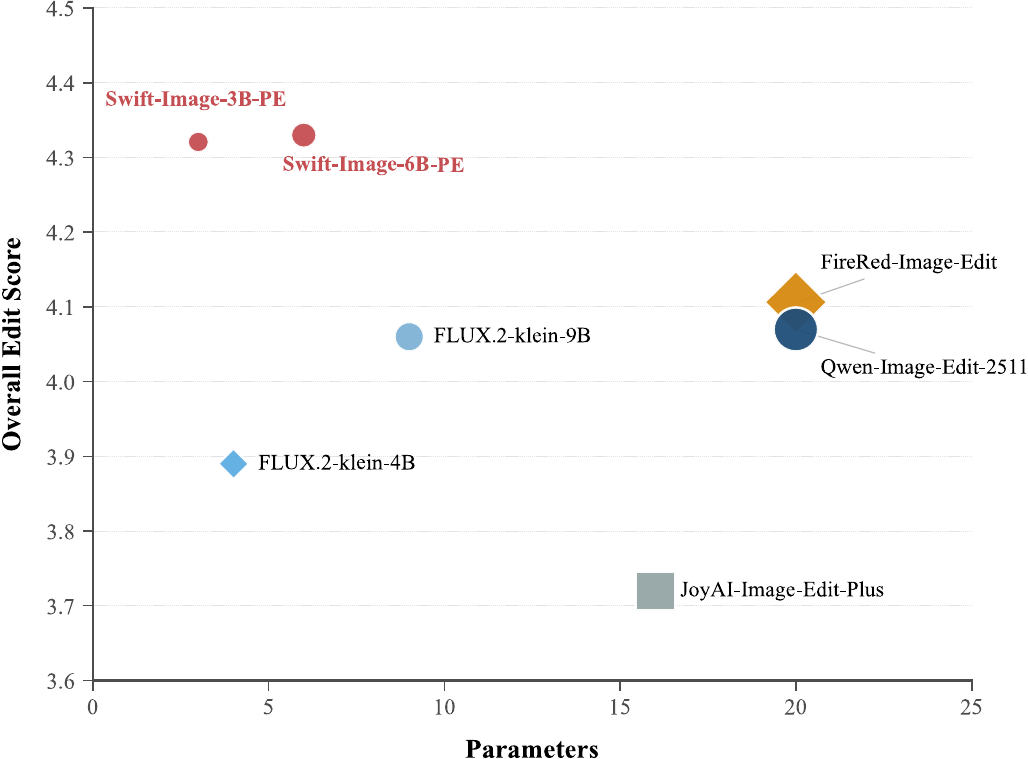}
  \caption{Comparison for parameter and overall edit score with open-source models. The overall edit score is computed as the average across three public editing benchmarks and our benchmarks. \ours achieves leading performance with only 6B and 3B parameters. }
  \vspace{-5mm}
  \label{fig:radar_comparison}
\end{figure}
Many techniques relevant to this problem have been explored in prior systems. Multimodal condition encoders and hybrid or single-stream Diffusion Transformers (DiTs) have become increasingly common in recent image generation systems \citep{yang2025qwen3, wu2025qwen, blackforest2025flux2klein, cao2025hunyuanimage, cai2025z, ai2025ming}. Progressive and multi-stage training has been explored to efficiently scale image generation models toward higher resolutions and quality \citep{zhao2026qwen, chen2026booguimage01}, while recent unified systems increasingly optimize generation and editing within closely shared architectures \citep{hidreamolimage, blackforest2025flux2klein}. Reinforcement-learning-based post-training has been applied to directly optimize image generation models toward downstream reward objectives \citep{xue2025dancegrpo, liu2025flow, zheng2025diffusionnft}, and prompt rewriting has been used to decouple high-level instruction interpretation from the underlying image generator \citep{wang2025promptenhancer, feng2026gen}. Model few-step distribution distillation provide complementary approaches for reducing deployment cost \citep{liu2025decoupled, yin2024improved, liu2026continuoustimedistributionmatchingfewstep}. We do not treat these components as novel in isolation. Instead, this work focuses on training engineering: how to organize and combine them into a coherent pipeline that produces a strong compact model. In practice, the effectiveness of such a system depends not only on the choice of individual components, but also on their interactions across architecture design, pre-training, post-training, prompt processing, and deployment-oriented compression.


We present \ours, a compact unified model for text-to-image generation, single-image editing, and multi-image editing. Its visual renderer is a \textbf{6B} parallel single-stream DiT conditioned on multimodal representations from a vision-language encoder \citep{yang2025qwen3, cai2025z, blackforest2025flux2klein}. The architecture adopts block-shared timestep modulation, parallel attention and MLP computation \citep{dehghani2023scaling, blackforest2025flux2klein}, 4D rotary positional encoding\citep{blackforest2025flux2klein}, and a unified representation of text and image conditions. Character-level tokenization\citep{team2025longcat} is applied to text intended to appear in generated images, while multi-image positional offsets and image-preceding input formatting support reference-conditioned editing. Together, these choices provide a single generative backbone for multiple generation and editing settings without task-specific model weights.

The model is optimized through a progressive training pipeline. Training begins with low-resolution text-to-image data to establish broad visual--semantic correspondence, and subsequently introduces higher resolutions, structurally complex content, and image-editing supervision, following the general principle of progressively staged optimization used in large-scale image generation. Continual pre-training shifts the model toward cleaner and higher-resolution data, followed by supervised fine-tuning on a smaller curated corpus. Resolution bucketing and task-routed sequence packing are used to provide native support for image generation with arbitrary aspect ratios and prompts of unrestricted length\citep{dehghani2023patch}. The resulting model is further post-trained with reward signals covering text--image alignment, visual quality, aesthetics, instruction following, reference consistency, facial identity, and text rendering, building upon recent progress in reward-based optimization of diffusion and flow-based image generators \citep{zheng2025diffusionnft, liu2025flow}. General and task-focused policies are then consolidated into the final unified model by multi-teacher on-policy distillation.

We additionally use a Prompt Enhancer (PE) to separate high-level request processing from pixel-level rendering. The PE translates short or complex user requests into more explicit visual descriptions, while the DiT focuses on realizing these descriptions. It is jointly trained for generation and editing, and is further optimized using both text-level rewrite feedback and image-level feedback obtained from the frozen renderer. Although PE--DiT decoupling is not new by itself \citep{wang2025promptenhancer}, it serves as a practical component of our compact-model pipeline. In our evaluations, PE consistently improves both the 3B and 6B models, with particularly large gains on knowledge-intensive, compositional, and layout-sensitive tasks.

For more efficient deployment, we structurally prune the 6B renderer and apply progressive recovery training and knowledge distillation to obtain a 3B variant. We further use distribution matching distillation to reduce the number of sampling steps \citep{liu2025decoupled, liu2026continuoustimedistributionmatchingfewstep}. These procedures produce a family of models with different trade-offs among parameter count, inference cost, and generation quality, while retaining the same unified task interface.

We evaluate \ours on benchmarks covering text-to-image generation, general image editing, single- and multi-image editing, practical applications, and knowledge-intensive visual tasks. The 6B model achieves leading aggregate performance among the evaluated open-source systems, while the compressed 3B model incurs \textbf{nearly no loss} at a smaller backbone scale. Reinforcement learning delivers substantial and consistent improvements over the reported editing benchmarks, and remarkably, the few-step distilled variant achieves even stronger overall performance with substantially fewer sampling steps.. The main 6B renderer training process consumes approximately \textbf{only 243K GPU hours}. These results demonstrate that a carefully assembled training recipe can produce strong and broadly applicable image generation and editing capabilities without relying on a tens-of-billions-parameter generative backbone.

\begin{figure*}[p]
    \centering
    \includegraphics[width=.96\linewidth]{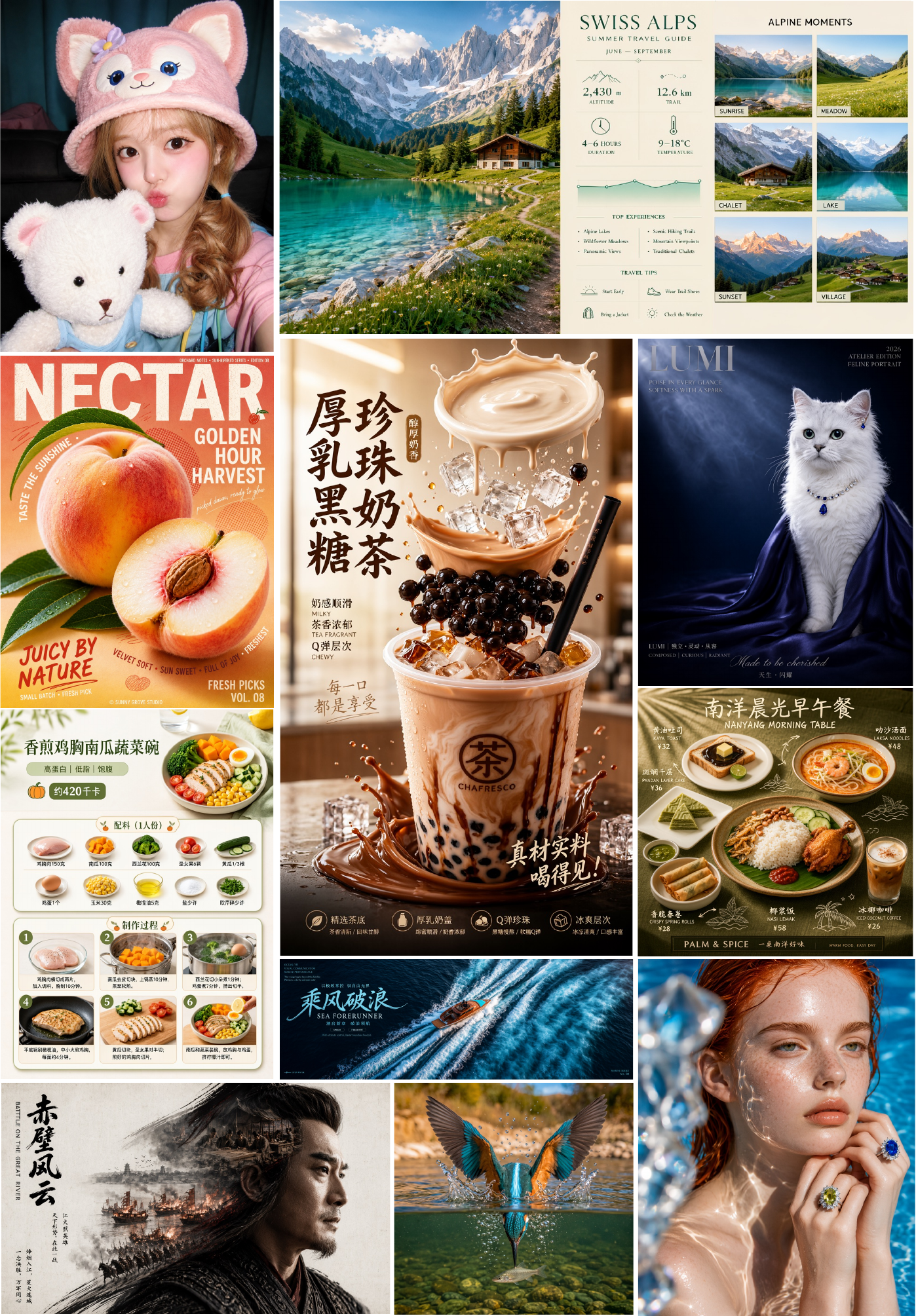}
    \caption{\ours generate diverse, high-fidelity images across distinct visual domains, including general images with photography-level realism, and posters with complex structure.}
    \label{fig:t2i_teaser}
\end{figure*}

\begin{figure*}[h]
    \centering
    \includegraphics[width=\linewidth]{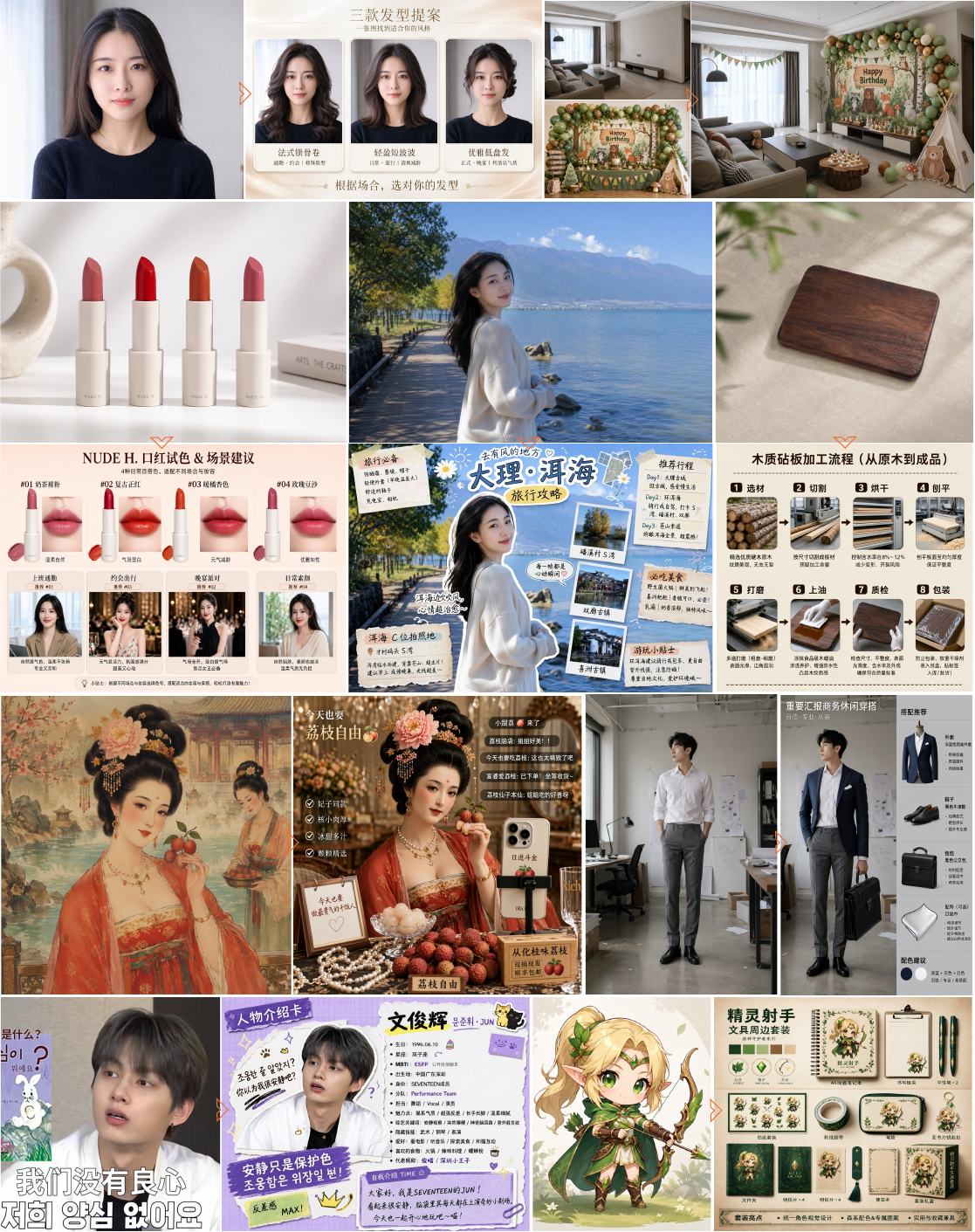}
    \caption{ Demonstration of \ours's complex editing capabilities. Moving beyond simple image modifications, the model demonstrates remarkable intelligence in handling tasks with high information density. From transforming a simple portrait into a comprehensive hairstyle guide and converting raw materials into industrial process charts, to generating text-intensive e-commerce advertisements and scene-consistent cinematic posters, \ours seamlessly integrates rich typography, multi-element layout formatting, and precise semantic editing under a unified framework.}
    \label{fig:i2i_reason_teaser2}
\end{figure*}

\begin{figure*}[h]
    \centering
    \includegraphics[width=.93\linewidth]{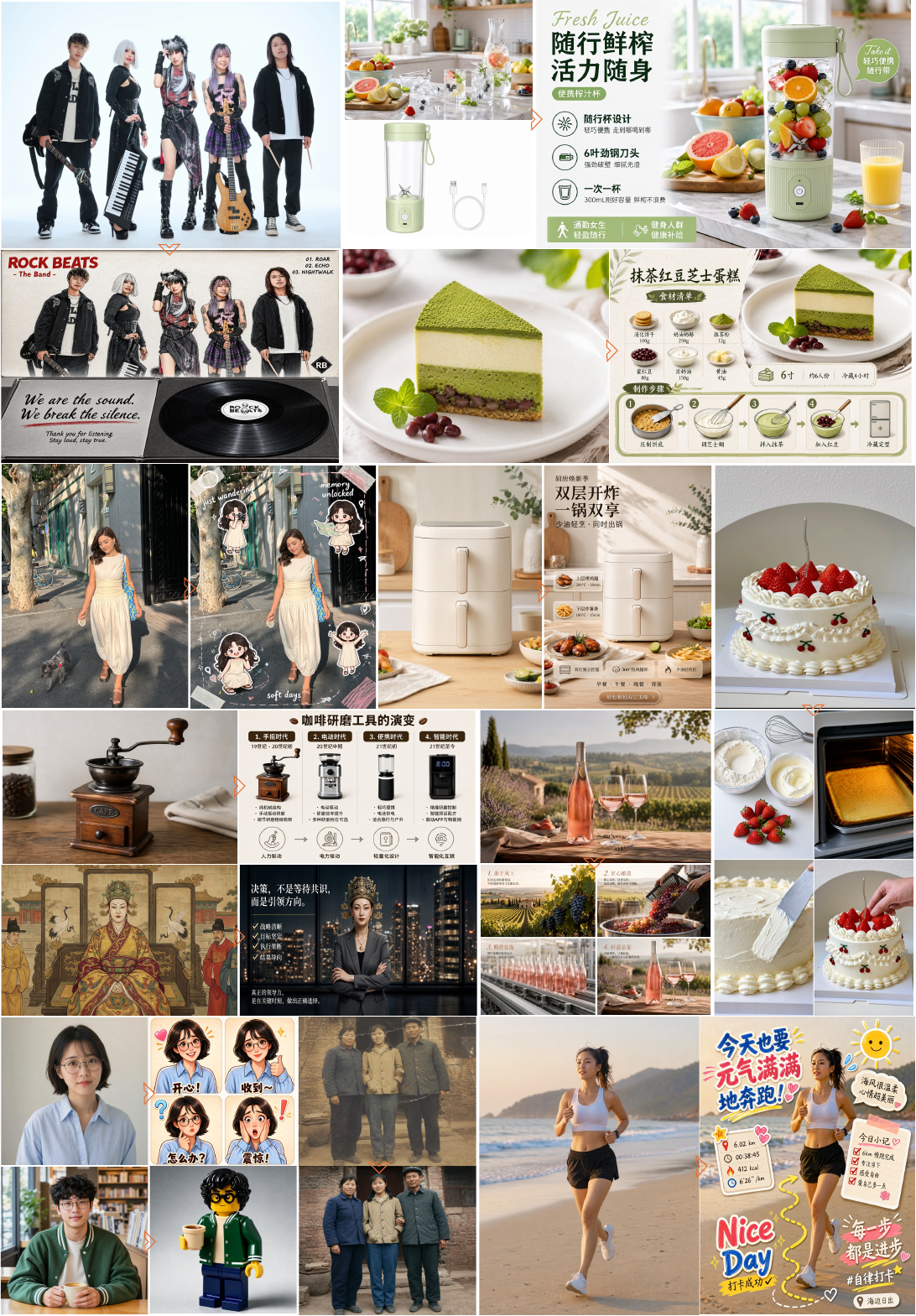}
    \caption{Extensive real-world applications of \ours, showcasing its mastery in text-intensive visual editing, including infographics, product advertisements, and storyboards.}
    \label{fig:i2i_reason_teaser3}
\end{figure*}

\begin{figure*}[t]
    \centering
    \includegraphics[width=1\linewidth]{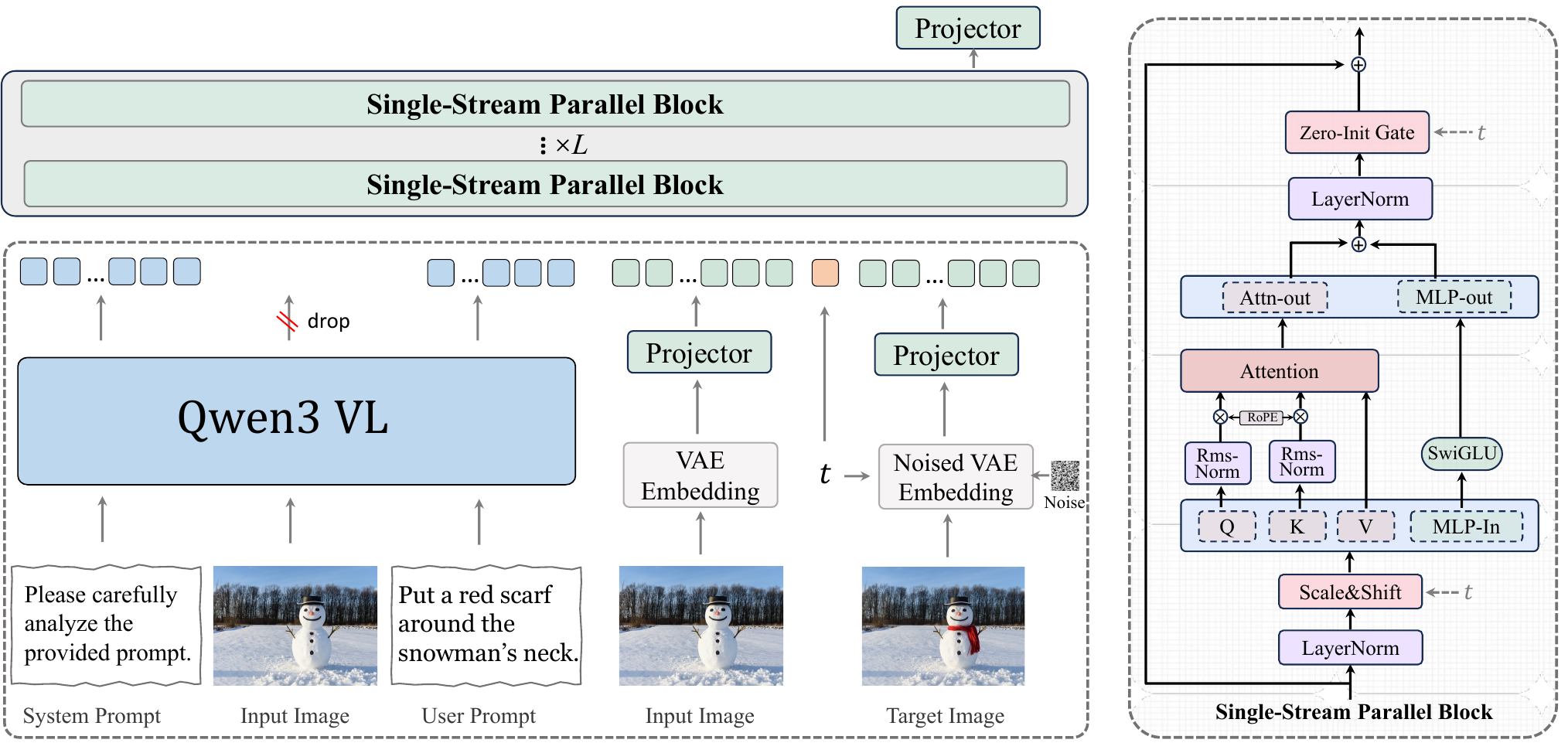}
    \caption{Unified visual backbone for T2I generation and image editing. Semantic conditions from Qwen3-VL and image latents from FLUX.2 AE are jointly processed by parallel single-stream Transformer blocks.}
    \label{fig:arch}
\end{figure*}

Our contributions are summarized as follows:
\begin{itemize}[leftmargin=*]
\item 
\textbf{Compact unified model.} We develop \ours, a compact unified model family supporting text-to-image generation, single-image editing, and multi-image editing. The \textbf{6B} model achieves leading aggregate performance among evaluated open-source systems with \textbf{only 243K GPU training hours}, while the compressed \textbf{3B} and few-step variants incur nearly no loss at substantially lower deployment cost (Figure \ref{fig:radar_comparison}, \ref{fig:t2i_teaser}, \ref{fig:i2i_reason_teaser2}, \ref{fig:i2i_reason_teaser3} and Table \ref{tab:editing_capability}, \ref{tab:eval_dimension} ).
\item 
\textbf{Systematic training recipe.} We establish an end-to-end training pipeline for compact unified image models, spanning capability-oriented progressive training, parallel expert RL, multi-teacher OPD, structural pruning, and few-step distillation. The pipeline effectively coordinates heterogeneous generation and editing objectives under limited model capacity and computation (Figure \ref{fig:arch}, \ref{fig:train_pipe}, \ref{fig:training_data_pyramid}, \ref{fig:rl_pipeline}, \ref{fig:distill}, \ref{fig:rewriter_pipeline}).
\item
\textbf{Practical insights from large-scale training.} Through extensive ablations and training experiments, we identify a set of transferable practices for compact visual models, including evolving the data distribution with model capability, prioritizing semantic coverage before preference optimization, specializing conflicting objectives before consolidation, and aligning architecture choices with generative learnability, efficiency, and training stability (Appdedix \ref{sec:appendix}).
\end{itemize}

\section{Related Work}
\subsection{Unified Image Generation and Editing}
Recent image generation systems are increasingly moving toward unified generation and editing with stronger multimodal conditioning. Qwen-Image~\cite{wu2025qwen}, Qwen-Image-2.0~\cite{zhao2026qwen}, and JoyAI-Image~\cite{song2026joyai} employ powerful MLLMs as condition encoders to improve instruction understanding, text rendering, and reference-conditioned generation, while models such as Krea-2 ~\cite{krea-2-2026} and Z-Image~\cite{cai2025z} explore efficient single-stream diffusion transformers. Recent systems further extend this paradigm toward practical unified generation and editing, including HiDream-O1-Image~\cite{hidreamolimage} and QwenImage 2.0~\cite{zhao2026qwen}. Reference-driven editing has also been explored through domain-specific systems study manipulable and commercial-scale virtual try-on~\cite{chen2024wear,chen2026tstars}. Nevertheless, high-performing open models often rely on large backbones or separate generation and editing variants. In contrast, \ours focuses on pushing the performance frontier of a compact unified model, jointly supporting text-to-image generation, single-image editing, and multi-image editing with one generative backbone.

\subsection{RL and Distillation for Diffusion Models}
Reinforcement learning has recently become an important post-training paradigm for diffusion and flow-based generative models. Flow-GRPO~\cite{liu2025flow} and DanceGRPO~\cite{xue2025dancegrpo} adapt GRPO-style optimization to modern visual generators, while DiffusionNFT~\cite{zheng2025diffusionnft} performs efficient online reward optimization through the diffusion forward process. Beyond single-policy RL, recent works investigate capability composition through on-policy distillation: DiffusionOPD~\cite{li2026diffusionopd} independently optimizes task-specific teachers and distills them into a unified student, while DanceOPD~\cite{zhou2026danceopd} formulates different generation and editing capabilities as expert generative fields. For inference acceleration, DMD-based approaches and Decoupled-DMD~\cite{liu2025decoupled} provide effective few-step distillation. \ours builds on these advances with parallel expert RL and multi-teacher OPD to mitigate cross-task interference, together with structural pruning and distribution matching distillation for compact and few-step deployment.

\subsection{Prompt Enhancement and Agentic Visual Reasoning}
As image generators become stronger, recent research increasingly separates high-level intent understanding from low-level visual rendering. Beyond conventional prompt rewriting~\cite{wang2025promptenhancer}, emerging approaches introduce explicit reasoning and external knowledge acquisition before generation. Gen-Searcher~\cite{feng2026gen} trains a multimodal agent with SFT and reinforcement learning to perform multi-hop web search, evidence reasoning, and visual-reference retrieval for knowledge-intensive image generation. Unify-Agent~\cite{chen2026unify} further formulates world-grounded synthesis as a unified process of prompt understanding, multimodal evidence searching, grounded recaptioning, and image generation. RS-Gen~\cite{bian2026rs} explores a training-free agentic framework that autonomously identifies reasoning and knowledge gaps and invokes search to resolve them. These works demonstrate the growing importance of reasoning before visual synthesis. ExpertVerse~\cite{wang2026expertversegeneralpurposebenchmarkexpertlevel} further benchmarks knowledge-intensive reasoning in visual synthesis. \ours follows a complementary direction: its Prompt Enhancer internalizes intent interpretation, knowledge reasoning, and layout planning into a learned rewriter, while the DiT is dedicated to precise rendering. The PE is further optimized with both rewrite-level and rendered-image rewards, enabling adaptive reasoning without requiring online search or tool invocation during normal generation.

\begin{figure*}[t]
    \centering
    \includegraphics[width=1\linewidth]{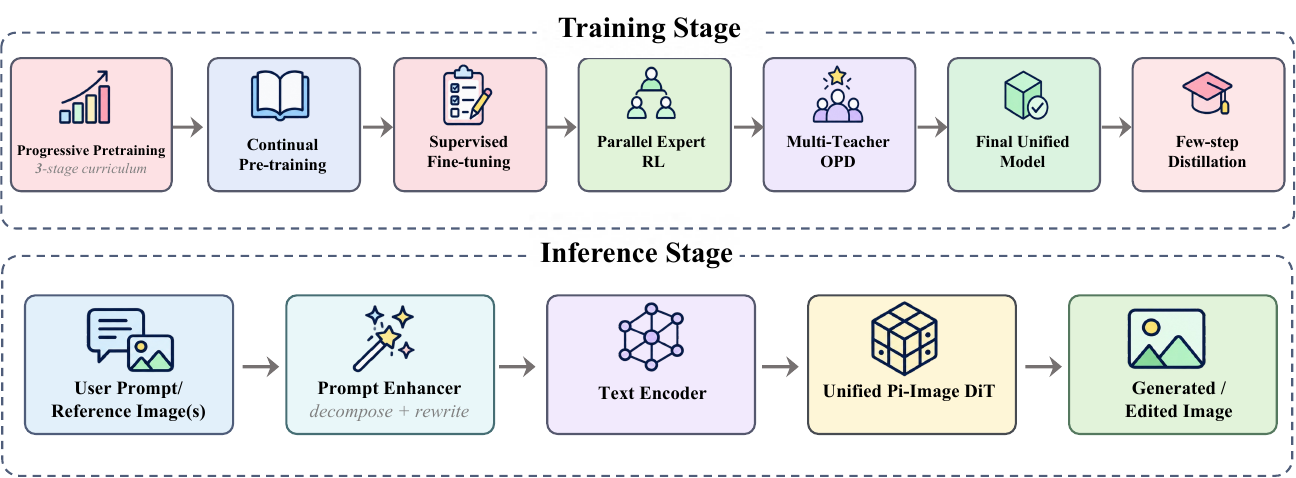}
    \caption{Method overview. The unified visual backbone undergoes progressive training, parallel expert RL, multi-teacher OPD, and deployment-oriented compression; at inference time, the Prompt Enhancer supplies a renderer-aligned visual specification.}
    \label{fig:train_pipe}
\end{figure*}

\section{Method}
\label{sec:method}
Our method separates intent understanding from visual realization while retaining a single visual backbone for text-to-image (T2I) generation and single- or multi-image editing (I2I). The Prompt Enhancer first converts a user request into an explicit visual specification when needed; a unified multimodal diffusion model then renders that specification. The visual model is learned through progressive pre-training and supervised fine-tuning (SFT), specialized with parallel expert reinforcement learning, consolidated by multi-teacher on-policy distillation (OPD), and compressed for parameter- and sampling-efficient deployment. Figure~\ref{fig:train_pipe} summarizes this pipeline.

\subsection{Unified Multimodal Diffusion Backbone}
\label{sec:architecture}

\paragraph{Multimodal conditioning.}
As shown in Figure~\ref{fig:arch}, a shared Qwen3-VL-8B encoder receives a System Prompt, zero or more input images, and the user instruction. Editing examples use an image-first ordering, and quoted strings that must be rendered in the output use character-level tokenization. We concatenate representations from three uniformly spaced hidden layers and map them into the diffusion hidden space with an MLP. The VLM's output image tokens are not forwarded to the DiT: they add sequence length without measurable quality gains in our ablations. Pixel-level reference information is nevertheless preserved through a separate path, where the input images and noised target image are encoded by FLUX.2 AE and linearly projected. A single System Prompt and the same conditioning interface are used across T2I and I2I.

\paragraph{Efficient parallel backbone.}
The conditional features and image latents are concatenated and processed by $L$ single-stream parallel blocks. Each block evaluates Multi-Head Attention and a SwiGLU MLP with expansion factor 3 in parallel. Omitting timestep modulation and normalization for clarity, the update is
\begin{equation}
    y' = \operatorname{LayerNorm}(x), \qquad
    y = x + \operatorname{MLP}(y') + \operatorname{Attention}(y').
    \label{eq:parallel_block}
\end{equation}
This structure permits operator fusion between the QKV and first MLP projections and between the attention-output and second MLP projections. Relative to the non-parallel alternative, it incurs only a minor quality difference while improving inference efficiency by approximately $10\%$. Timestep-dependent scale, shift, and zero-initialized gates are produced by one block-shared modulation network. Although sharing slightly weakens the modulation module in isolation, it frees a substantial parameter budget for attention and feed-forward layers and improves performance at a fixed model size.

\paragraph{Positioning and optimization stability.}
We use 4D-RoPE over \texttt{[T, H, W, L]}. Text occupies only the language axis $L$, while consecutive input images are separated by an offset of 10 along $T$. The resulting image-index prior prevents tokens at identical spatial coordinates in different images from becoming artificially close and reduces copy-paste artifacts in multi-image editing~\citep{xia2025dreamomni2}. For stable optimization, we retain standard residual connections, apply RMSNorm only to QK, use LayerNorm elsewhere, place Sandwich Norm around attention and feed-forward branches, and constrain modulation gates with Tanh~\citep{zhuo2024lumina,cai2025z}. All projection and modulation layers are bias-free.

\paragraph{Design insights.}
Our comparisons reveal two broader lessons. First, stronger multimodal-understanding benchmarks do not guarantee better generative conditioning: Qwen3.5-4B~\citep{qwen3.5} converges more slowly than Qwen3-VL-8B~\citep{yang2025qwen3} in our diffusion training. Second, early convergence alone is insufficient for choosing a latent encoder. Our internal PAE~\citep{yue2026matters} learns quickly at first, but reconstruction errors introduce text and image artifacts that constrain the final quality ceiling; FLUX.2 AE~\citep{blackforest2025flux2klein} provides the best overall reconstruction--convergence trade-off. The complete design space, component-level findings, negative results, and optimizer study are reported in Appendix~\ref{app:architecture} and Table~\ref{tab:ablation_summary}.

\subsection{Progressive Unified Training}
\label{sec:progressive_training}

We train the visual backbone with a coarse-to-fine curriculum along three axes: spatial resolution, task complexity, and data quality. The central principle is to increase resolution before introducing heterogeneous editing objectives, avoiding a simultaneous shift in spatial scale and task distribution. The underlying training corpus follows the capability-centric data design of \textit{Wang et al.}\cite{wang2026corpora}, where heterogeneous generation and editing supervision is constructed and organized according to the dependency structure of generative capabilities.
\paragraph{Foundational pre-training.}
The first 500K steps prioritize scale, diversity, semantic coverage, text--image alignment, and structural modeling. Training begins with low-resolution T2I, raises the resolution from 256px to 512px, and only then introduces joint T2I and editing data. This progression moves the model from broad visual semantics to fine-grained, unified generation and editing. Aspect-ratio buckets reduce padding and cropping, while task-routed \textbf{sequence packing}\cite{dehghani2023patch} prevents the short prompts common in T2I from being batched inefficiently with the very long conditions found in dense text rendering and complex editing, and natively supports generation with \textbf{unlimited resolution and prompt length}.

\paragraph{Continual pre-training and SFT.}
Continual pre-training (CT) runs for 200K steps while increasing resolution from 512px to 1024px and shifting the data distribution toward visually stronger sources. VLM-based hierarchical semantic labels and category-level resampling counteract source bias and retain long-tail coverage; editing data receive additional instruction-category balancing. SFT then runs for 10K steps at 1024px on a compact, human-verified high-quality subset, with evaluation-driven augmentation of weak capabilities. We use Logit-Normal timestep sampling during pre-training to emphasize medium-noise states and global structure, then switch to Uniform Sampling during SFT to distribute supervision across the trajectory and strengthen fine textures. Figure~\ref{fig:training_data_pyramid}, Table~\ref{tab:training_configurations}, and Appendix~\ref{app:progressive_training} provide the complete data flow, packing scheme, stage-wise filtering, and hyperparameters.

\subsection{Parallel Reinforcement Learning and Multi-Expert Consolidation}
\label{sec:expert_post_training}

Supervised training endows the model with broad generation and editing capabilities, but does not explicitly optimize human preferences such as aesthetics, realism, reference consistency, and fine-grained instruction following. Recent studies have therefore adopted reinforcement learning methods\cite{liu2025flow,zheng2025diffusionnft} to further push the performance frontier. However, scaling RL to heterogeneous multi-task data remains challenging. Conflicting task objectives induce gradient interference, resulting in seesaw trade-offs across task categories. Meanwhile, the limited sample efficiency of group-rollout RL requires extended optimization for under-optimized tasks, whereas prolonged diffusion RL is prone to instability and mode collapse. Consequently, some tasks fail to reach their task-specific performance ceilings under mixed-task training. We address both problems with a two-stage procedure: \textbf{task-coherent policies are optimized in parallel and then merged by multi-teacher on-policy distillation (OPD)~\citep{li2026diffusionopd,zhou2026danceopd}.}

\paragraph{Multi-dimensional Reward System.}
We construct an online multi-dimensional reward system to provide comprehensive and discriminative feedback across generation tasks. For text-to-image (T2I), the rubric-based evaluator assesses text--image alignment, category-aware aesthetics, visual quality, and style consistency. For image editing, it evaluates instruction following, reference-image consistency, and visual quality. We further incorporate auxiliary rewards based on ArcFace~\citep{deng2019arcface} and PP-OCRv6~\citep{zhang2026ppocrv6} to measure identity preservation and text-rendering accuracy, respectively.

\paragraph{Guidance-enhanced Rollout.}
We enable classifier-free guidance (CFG) during rollout to improve the quality of sampled trajectories and provide stronger candidates for reward evaluation. Policy optimization is performed without CFG to reduce computational overhead and improve training efficiency. This asymmetric design retains the benefits of guidance-enhanced reward signals while implicitly transferring part of the guided sampling capability to the unguided policy, without incurring the full cost of guided optimization.

\paragraph{Task-aware Reward Routing and Parallel Experts.}
Because different task categories require distinct optimization objectives, we flexibly route and compose rewards from the multi-dimensional reward pool according to the task type. Rather than forcing heterogeneous and potentially conflicting objectives through a single policy, we employ DiffusionNFT~\citep{zheng2025diffusionnft} to train multiple task-coherent policies in parallel, including a dedicated T2I expert, a general-purpose editing expert, and several domain-specific editing experts targeting subdomains that remain under-optimized during joint training. This decomposition mitigates cross-task reward and gradient interference while preserving broad task coverage and strong domain-specific performance.

\paragraph{Full-step multi-teacher OPD.}
The experts' improvements initially reside in separate checkpoints. We initialize a shared student from the mixed-task policy and distill each on-policy sample from its domain-matched expert, combining generalist balance with specialist strength. Beyond this stage, we further apply MOPD to post-pruning 6B-to-3B transfer and task-specific few-step consolidation for sampling acceleration. The full reward definitions, expert routing, and RL/OPD workflow are detailed in Appendix~\ref{app:post_training} and Figure~\ref{fig:rl_pipeline}.



\subsection{Deployment-Oriented Compression}
\label{sec:compression}
We compress the unified model along two complementary dimensions: sampling steps and parameter count. Distribution matching reduces the denoising trajectory from 50 steps to 8, and structural pruning yields a deployable 3B backbone.

\subsubsection{Few-Step Distillation}
\label{sec:few_step_distillation}

Our few-step recipe builds on Distribution Matching Distillation (DMD/DMD2)~\citep{yin2024one,yin2024improved}: a frozen teacher supplies the real score, a jointly trained fake-score network models the student distribution, and their difference supplies the distribution-matching gradient. To reduce the exposure bias between training states and the student's inference trajectory, we draw the intermediate states from the student's own sampling trajectory rather than from forward-diffused training data. Following Continuous-Time Distribution Matching (CDM)~\citep{liu2026continuoustimedistributionmatchingfewstep}, \textbf{Dynamic Backward Simulation} samples a step budget from $\{2,4,8,16,28\}$ and partially unrolls the student from noise, exposing it to self-generated states of varying quality.

\begin{figure*}[t]
  \includegraphics[width=1.0\textwidth]{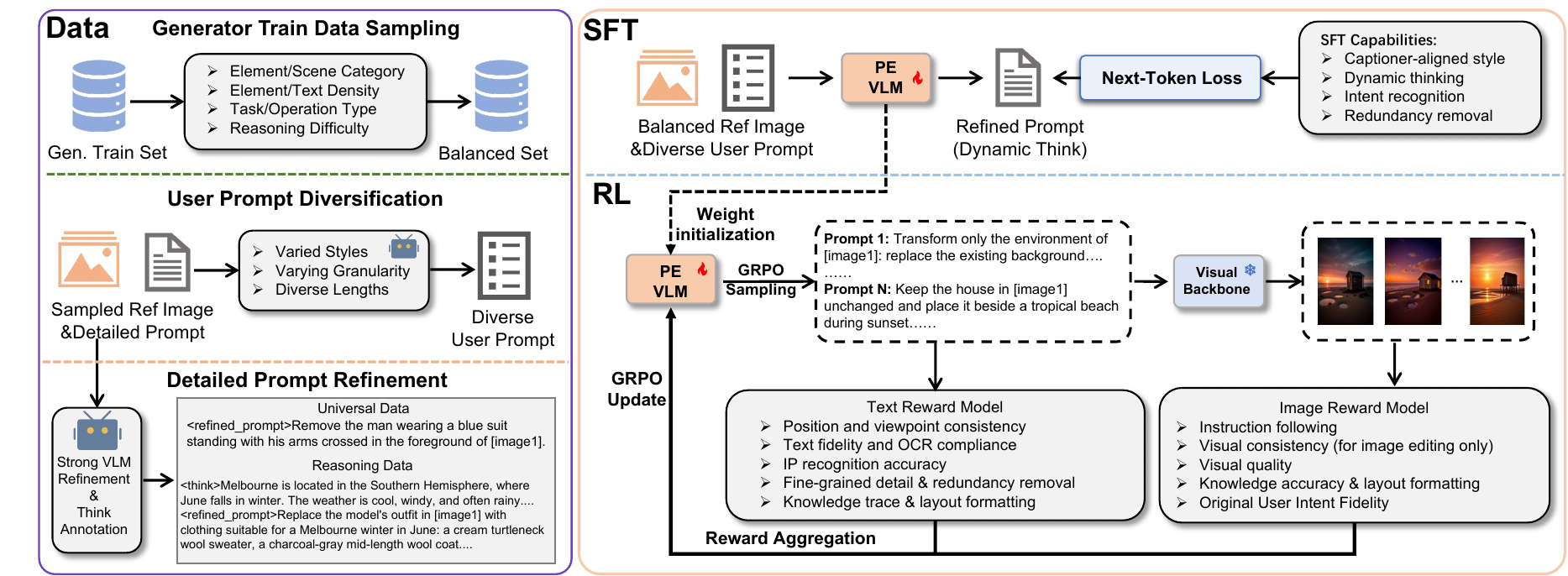}
  \caption{Overview of Prompt Enhancer data construction and training. Balanced and diversified T2I/I2I data are verified and reasoning-annotated by a teacher VLM. SFT establishes a unified output style and complexity-adaptive reasoning, after which renderer-grounded GRPO jointly optimizes textual correctness and visual executability using feedback from the frozen visual backbone.}
  \label{fig:rewriter_pipeline}
\end{figure*}
We make three stability-oriented changes. First, the stopping distribution favors later, lower-noise states, which concentrates supervision on texture and high-frequency refinement while retaining coverage of global structure. Second, following Decoupled-DMD~\citep{liu2025decoupled}, each predicted $x_0$ is re-noised only below the backward-simulation stopping noise, so the teacher does not supervise noisier states the student has already traversed. Third, an adversarial head is attached to a late layer of the frozen teacher instead of the continually changing fake score, providing a fixed semantic feature space and a lower-variance signal. In heterogeneous 6B-to-3B distillation, the fake score adopts the 3B student architecture, whereas the discriminator remains anchored to the 6B teacher.

One teacher does not optimally balance all tasks: T2I-oriented supervision favors generation quality, while editing-oriented supervision better preserves reference consistency. We therefore distill separate few-step T2I and editing experts, consolidate them with a second, few-step multi-teacher OPD stage~\citep{li2026diffusionopd,zhou2026danceopd}, and finish with\textbf{ mixed-task few-step DiffusionNFT}~\citep{zheng2025diffusionnft}. The complete procedure is shown in Figure~\ref{fig:distill} and detailed in Appendix~\ref{app:compression}.

\subsubsection{Structured Model Pruning}
\label{sec:pruning}

The 3B model is obtained from the 6B model by reducing the number of attention heads from 32 to 24 and directly inheriting the surviving parameters. Instead of applying a short recovery stage, the student re-enters the coarse-to-fine training curriculum through CT. During SFT, a frozen 6B teacher receives the same noisy latent $z_t$, timestep $t$, and condition $c$, and supervises the student's velocity prediction:
\begin{equation}
    \mathcal{L}_{\mathrm{KD}}
    =
    \left\|
    v_{\mathrm{3B}}(z_t,t,c)
    -
    \operatorname{sg}\!\left(v_{\mathrm{6B}}(z_t,t,c)\right)
    \right\|_2^2,
    \label{eq:compact_kd}
\end{equation}
\begin{equation}
    \mathcal{L}
    =
    \mathcal{L}_{\mathrm{diff}}
    +
    \lambda_{\mathrm{KD}}\mathcal{L}_{\mathrm{KD}},
    \label{eq:compact_objective}
\end{equation}
where $\operatorname{sg}$ stops gradients, $\mathcal{L}_{\mathrm{diff}}$ is the original diffusion objective, and $\lambda_{\mathrm{KD}}$ controls teacher supervision. Finally, the SFT-trained 3B model becomes the student in cross-capacity OPD, with the specialized \textbf{6B RL policies acting as task-routed teachers}. This sequence restores capabilities lost to pruning and transfers complementary generation and editing improvements into one compact model.

\subsection{Reasoning Before Rendering: Prompt Enhancer}
\label{sec:pe}
The Prompt Enhancer (PE) implements our separation between high-level reasoning and pixel-level rendering. Given a user instruction and optional reference images, it produces a structured \texttt{refined\_prompt} aligned with the visual backbone~\citep{wang2025promptenhancer}. Straightforward requests are rewritten directly; ambiguous, knowledge-intensive, or layout-sensitive requests first trigger a \texttt{<think>} trace. Only the final \texttt{refined\_prompt} is passed to the DiT, so the renderer receives an explicit visual specification without also having to learn the PE's deliberation process. A single PE handles both T2I and I2I and allocates extra computation only when the request requires it.



\paragraph{Reasoning-aware supervision and SFT.}
We construct two complementary data sources. \emph{Universal data} are derived from the visual descriptions produced by the captioning pipeline used to train our backbone. A strong teacher first verifies and, when necessary, completes each captioner-produced refined prompt, and then supplements it with an appropriate reasoning trace. Corresponding user inputs are synthesized in diverse lengths, linguistic styles, and structured formats, and are further mixed with real user queries. \emph{Reasoning data} target requests that require intent disambiguation, domain knowledge, or spatial planning. Their desired outcomes are first rendered, after which a teacher generates a refined prompt conditioned on the realized target. This outcome-grounded supervision encodes a solved visual plan rather than a surface expansion of the original request; target images are used only during offline data construction and are unavailable to the PE at inference time.
\begin{table*}[t]
    \centering
    \caption{
    Comparison of image editing models across three public benchmarks and ourproposed Pi-Benchmark suite. Best and second-best results among open-source models are shown in \textbf{bold} and \underline{underlined}, respectively. For multilingual benchmarks, we report results on the English split.
    }
    \label{tab:editing_capability}

    \resizebox{0.99\linewidth}{!}{
    \begin{threeparttable}
    \small
    \setlength{\tabcolsep}{3pt}
    \renewcommand{\arraystretch}{1.2}

    \begin{tabular}{l| c c c c c c c c}
        \toprule
        \textbf{Model}
        & \textbf{Params.}
        & \textbf{GEdit}
        & \textbf{ImgEdit}
        & \textbf{REDEdit}
        & \textbf{CPI-General}
        & \textbf{CPI-Practical}
        & \textbf{Overall$\uparrow$}$^{\dagger}$
        & \textbf{Rank} \\
        \midrule
        \rowcolor{groupgray}[0pt][0pt]
        \multicolumn{9}{@{}l@{}}{
            \hspace{0.5em}\textit{\textbf{Proprietary / Closed-source Models}}
        } \\

        GPT-Image-2\cite{gptimage2_model_card}
        & --
        & 8.69
        & 4.74
        & 4.65
        & 4.64
        & 4.69
        & 4.61
        & 1
        \\

        Nano Banana Pro\cite{google2025nanobanana}
        & --
        & 7.73
        & 4.37
        & 4.42
        & 4.47
        & 4.58
        & 4.34
        & 6
        \\

        Seedream5 Pro\cite{seedream2025seedream}
        & --
        & 8.63
        & 4.57
        & 4.62
        & 4.63
        & 4.72
        & 4.57
        & 2
        \\
        
        Seedream4.5\cite{seedream2025seedream}
        & --
        & 7.82
        & 4.32
        & 4.20
        & 4.32
        & 4.33
        & 4.22
        & 9
        \\

        Qwen Image 2.0 Pro\cite{zhao2026qwen}
        & --
        & 8.52
        & 4.45
        & 4.38
        & 4.38
        & 4.39
        & 4.37
        & 5 \\
        \midrule
        \rowcolor{groupgray}[0pt][0pt]
        \multicolumn{9}{@{}l@{}}{
            \hspace{0.5em}\textit{\textbf{Open-source Models}}
        } \\

        Qwen-Image-Edit-2511~\cite{wu2025qwen}
        & 20B
        & 7.87
        & 4.51
        & 4.23
        & 3.73
        & 3.85
        & 4.07
        & 13
        \\

        LongCat-Image-Edit~\cite{team2025longcat}
        & 6B
        & 7.74
        & 4.45
        & 4.12
        & -
        & -
        & -
        & -
        \\
        
        FLUX.2-klein-9B~\cite{blackforest2025flux2klein}
        & 9B
        & 8.12
        & 4.32
        & 4.07
        & 3.78
        & 3.86
        & 4.05
        & 14
        \\
        
        FLUX.2-klein-4B~\cite{blackforest2025flux2klein}
        & 4B
        & 7.79
        & 4.12
        & 3.94
        & 3.65
        & 3.63
        & 3.88
        & 15
        \\
        
        FireRed-Image-Edit~\cite{firered2026rededit}
        & 20B
        & 7.94
        & 4.56
        & 4.26
        & 3.76
        & 3.89
        & 4.10
        & 12
        \\

        JoyAI-Image-Edit~\cite{song2026joyai}
        & 16B
        & 8.27
        & 4.46
        & 4.17
        & -
        & -
        & -
        & -
        \\
        
        JoyAI-Image-Edit-Plus~\cite{song2026joyai}
        & 16B
        & 7.54
        & 4.02
        & 3.87
        & 3.32
        & 3.50
        & 3.72
        & 16
        \\
        
        Boogu-Image-0.1-Edit~\cite{chen2026booguimage01}
        & 10B
        & 7.95
        & 4.51
        & 4.01
        & -
        & -
        & -
        & -
        \\

        Boogu-Image-0.1-Edit-Thinking~\cite{chen2026booguimage01}
        & 10B
        & 8.24
        & \underline{4.64}
        & 4.11
        & -
        & -
        & - 
        & -
        \\
        
        \ours-3B
        & 3B
        & 8.10
        & 4.56
        & 4.29
        & 3.95
        & 3.91
        & 4.15
        & 11
        \\
        
        \ours-3B w/ PE
        & 3B
        & 8.17
        & 4.58
        & 4.37
        & 4.27
        & 4.23
        & 4.31
        & 8
        \\

        \ours-3B w/ PE (API)
        & 3B
        & \underline{8.34}
        & 4.60
        & \underline{4.41}
        & \underline{4.35}
        & \underline{4.47}
        & \underline{4.40}
        & 4
        \\
        
        \rowcolor{oursred}
        \textbf{\ours-6B}
        & 6B
        & 8.03
        & 4.56
        & 4.34
        & 3.96
        & 3.92
        & 4.16
        & 10
        \\

        \rowcolor{oursred}
        \textbf{\ours-6B w/ PE}
        & 6B
        & 8.32
        & 4.63
        & 4.39
        & 4.28
        & 4.19
        & 4.33
        & 7
        \\

        \rowcolor{oursred}
        \textbf{\ours-6B w/ PE (API)}
        & 6B
        & \textbf{8.35}
        & \textbf{4.64}
        & \textbf{4.45}
        & \textbf{4.35}
        & \textbf{4.47}
        & \textbf{4.41}
        & 3
        \\

        \bottomrule
    \end{tabular}

    \begin{tablenotes}[flushleft]
        \footnotesize
        \item[*] The parameter counts of proprietary models are not publicly disclosed. Proprietary models are evaluated through their official APIs.
    
        \item[$\dagger$] The overall score is the unweighted arithmetic mean of the five benchmark scores listed in the table. Since GEdit uses a 10-point scale while the other benchmarks use a 5-point scale, its score is divided by two before averaging. Overall scores and rankings are reported only for models evaluated on all five benchmarks. Rankings are calculated jointly across proprietary and open-source models based on
        the unrounded overall scores.
    \end{tablenotes}
    \end{threeparttable}}
 \vspace{-3mm}
\end{table*}
After balancing Universal/Reasoning, T2I/I2I, and fine-grained task categories, each example is assigned a reasoning mode based on complexity. Complex examples supervise both \texttt{<think>} and \texttt{refined\_prompt}; simple examples supervise only \texttt{refined\_prompt}. SFT therefore learns a common output space and an adaptive reasoning policy simultaneously. For T2I, the refined prompt gives a complete visual description. For I2I, unchanged content is referenced through \texttt{[image N]}, while requested modifications, newly introduced content, and target attributes are described explicitly at the same granularity as T2I.

\paragraph{Renderer-grounded reinforcement learning.}
Starting from the SFT checkpoint, we use Group Relative Policy Optimization (GRPO)~\citep{xue2025dancegrpo} while keeping the visual DiT frozen (Figure~\ref{fig:rewriter_pipeline}). For each request, the PE samples a group of rewrites with adaptive reasoning. A \emph{Rewrite Text Reward} measures text-level instruction following---including spatial relations, exact text-rendering constraints, and completeness---and knowledge reasoning, including entity recognition, factuality, and layout planning. In parallel, every refined prompt is rendered by the DiT. An \emph{Image Generation Reward} then measures whether the result realizes the requested content, knowledge, and layout, together with visual quality, aesthetics, and reference consistency for editing.

These views serve different roles: the text reward protects correctness and controllability, whereas the image reward tests whether the specification is executable by the particular renderer. We standardize each reward within its GRPO sampling group and fuse the resulting relative advantages instead of combining raw scores with hand-tuned scales. Only the PE is updated, allowing its phrasing, detail, and constraint granularity to adapt to the frozen DiT without changing the renderer. Appendix~\ref{app:prompt_enhancer} details teacher-based data synthesis, output conventions, balancing, and the full reward rubric; Figure~\ref{fig:pe_case} provides qualitative comparisons.

\section{Performance and Evaluation}
\subsection{Image Editing Evaluation}

\paragraph{Evaluation setup.}
We evaluate \ours on three widely adopted public benchmarks:
GEdit-Bench~\cite{liu2025step1x-edit},
ImgEdit-Bench~\cite{ye2025imgedit}, and
REDEdit-Bench~\cite{firered2026rededit}.
These benchmarks assess general editing quality, instruction following, and
content preservation. We further evaluate practical and reasoning-intensive
editing on CPI-Benchmark suite~\cite{zhou2026cpi}, including CPI-General, CPI-Practical, and CPI-Intelligent. Unless otherwise specified, proprietary models are
evaluated through their official APIs, and the English split is reported for
multilingual benchmarks.

\paragraph{Overall performance.}
As summarized in Table~\ref{tab:editing_capability}, \ours delivers strong
and balanced editing performance while using substantially fewer parameters
than most competing open-source editors. Across the three public benchmarks,
CPI-General, and CPI-Practical, \ours-6B with API-based PE achieves the highest
overall score among the evaluated open-source models and ranks third among
all evaluated systems. Meanwhile, \ours-3B with API-based PE achieves the
second-best open-source overall score despite having only 3B parameters.
These results demonstrate the effectiveness of our compact unified
generation--editing architecture and the complementary benefit of prompt
enhancement.
\begin{table}[t]
    \centering
    \caption{
    Results on CPI-Intelligent under standard and PE-enhanced prompting. 
    Best and second-best results among open-source models are shown in \textbf{bold} and \underline{underlined}, respectively.
    }
    \label{tab:i2i_reasoning}

    \footnotesize
    \setlength{\tabcolsep}{3pt}
    \renewcommand{\arraystretch}{1.2}
    \resizebox{\columnwidth}{!}{
    \begin{tabular}{l| c c c c c c c c c}
        \toprule
        \textbf{Model}
        & \textbf{IE}
        & \textbf{GH}
        & \textbf{FN}
        & \textbf{DI}
        & \textbf{FB}
        & \textbf{LC}
        & \textbf{SE}
        & \textbf{DP}
        & \textbf{Overall$\uparrow$} \\
        \midrule
        \rowcolor{groupgray}[0pt][0pt]
        \multicolumn{10}{@{}l@{}}{
            \hspace{0.5em}\textit{\textbf{Proprietary / Closed-source Models}}
        } \\

        GPT-Image-2\cite{gptimage2_model_card} & 5.00 & 4.47 & 4.86 & 4.76 & 4.82 & 4.92 & 4.38 & 4.93 & 4.77 \\
        Nano Banana Pro\cite{google2025nanobanana} & 4.84 & 4.59 & 4.58 & 4.50 & 4.81 & 4.76 & 4.37 & 4.87 & 4.68 \\
        Seedream5 Pro\cite{seedream2025seedream} & 4.83 & 4.63 & 4.75 & 4.67 & 4.68 & 4.80 & 4.26 & 4.92 & 4.70 \\
        Seedream4.5\cite{seedream2025seedream} & 4.31 & 3.47 & 4.35 & 2.92 & 4.23 & 3.98 & 3.12 & 3.96 & 3.79 \\
        Qwen Image 2.0 Pro\cite{zhao2026qwen} & 4.29 & 3.67 & 4.59 & 2.53 & 4.11 & 3.92 & 3.09 & 4.19 & 3.79 \\

        \midrule
        \rowcolor{groupgray}[0pt][0pt]
        \multicolumn{10}{@{}l@{}}{
            \hspace{0.5em}\textit{\textbf{Open-source Models}}
        } \\

        \multicolumn{10}{@{}l@{}}{
            \hspace{1.5em}\textcolor{gray}{\textit{Standard Evaluation (w/o PE)}}
        } \\
        Qwen-Image-Edit-2511~\cite{wu2025qwen} & 3.51 & 2.46 & 2.90 & 1.54 & 3.06 & 2.48 & 1.57 & 2.88 & 2.54 \\
        FLUX.2-klein-9B~\cite{blackforest2025flux2klein} & 3.57 & 2.04 & 2.74 & 1.71 & 2.98 & 2.59 & 1.65 & 2.67 & 2.48 \\
        FLUX.2-klein-4B~\cite{blackforest2025flux2klein} & 3.45 & 1.83 & 2.52 & 1.66 & 3.01 & 2.38 & 1.58 & 2.39 & 2.33 \\
        FireRed-Image-Edit-1.0~\cite{firered2026rededit} & 3.74 & 2.31 & 2.99 & 1.66 & 3.22 & 2.68 & 1.75 & 2.91 & 2.65 \\
        JoyAI-Image-Edit-Plus~\cite{song2026joyai} & 2.79 & 1.83 & 1.96 & 1.57 & 2.46 & 2.14 & 1.54 & 2.21 & 2.07 \\
        \ours-3B
        & 2.76
        & 1.78
        & 1.91
        & 1.55
        & 2.84
        & 1.86
        & 1.50
        & 2.04
        & 2.02 \\
        
        \rowcolor{oursred}
        \textbf{\ours-6B}
        & 3.05
        & 1.90
        & 1.97
        & 1.78
        & 2.91
        & 2.28
        & 1.63
        & 2.46
        & 2.26 \\

        \multicolumn{10}{@{}l@{}}{
            \hspace{1.5em}\textcolor{gray}{
                \textit{With Prompt Enhancer (w/ PE)}
            }
        } \\

        Qwen-Image-Edit-2511~\cite{wu2025qwen}
        & 4.33
        & 3.47
        & 3.77
        & 2.27
        & 3.80
        & 3.29
        & 2.60
        & 3.43
        & 3.36 \\

        FLUX.2-klein-9B~\cite{blackforest2025flux2klein}
        & 4.49
        & 3.30
        & 3.62
        & 2.16
        & 3.50
        & 3.43
        & 2.34
        & 3.36
        & 3.28 \\

        FLUX.2-klein-4B~\cite{blackforest2025flux2klein}
        & 4.18
        & 2.94
        & 3.72
        & 2.08
        & 3.47
        & 3.29
        & 2.08
        & 3.10
        & 3.10 \\

        FireRed-Image-Edit-1.0~\cite{firered2026rededit}
        & 4.30
        & 3.58
        & 3.60
        & 2.27
        & 3.78
        & 3.24
        & 2.71
        & 3.50
        & 3.37 \\

        JoyAI-Image-Edit-Plus~\cite{song2026joyai}
        & 4.63
        & 3.70
        & 4.33
        & 3.52
        & 4.28
        & 4.17
        & 3.37
        & \underline{4.38}
        & 4.04 \\

        \ours-3B
        & 4.60
        & 3.70
        & \underline{4.41}
        & 3.52
        & 4.50
        & 4.37
        & 3.56
        & 4.12
        & 4.10 \\

        \rowcolor{oursred}
        \textbf{\ours-6B}
        & 4.69
        & 3.76
        & \textbf{4.55}
        & \textbf{3.84}
        & \textbf{4.63}
        & \underline{4.38}
        & \textbf{3.70}
        & \textbf{4.38}
        & \underline{4.23} \\

        \multicolumn{10}{@{}l@{}}{
            \hspace{1.5em}\textcolor{gray}{
                \textit{With API-based Prompt Enhancer (w/ PE, API)}
            }
        } \\

        \ours-3B
        & \textbf{4.79}
        & \underline{4.03}
        & 4.07
        & 3.48
        & \underline{4.57}
        & 4.33
        & 3.28
        & 4.04
        & 4.10 \\

        \rowcolor{oursred}
        \textbf{\ours-6B}
        & \underline{4.78}
        & \textbf{4.14}
        & 4.38
        & \underline{3.59}
        & 4.54
        & \textbf{4.46}
        & \underline{3.70}
        & 4.24
        & \textbf{4.24} \\

        \bottomrule

    \end{tabular}}

    \parbox{\columnwidth}{%
        \scriptsize
        $^{*}$Abbreviations:
        IE = IP \& Entertainment,
        GH = Geography \& History,
        FN = Film \& Narrative,
        DI = Data \& Information,
        FB = Fashion \& Beauty,
        LC = Life \& Consumption,
        SE = Science \& Education,
        DP = Design \& Product.
    }
\end{table}
\paragraph{Public editing benchmarks.}
GEdit-Bench and ImgEdit-Bench evaluate general instruction-based editing,
with an emphasis on instruction adherence, visual quality, and preservation
of irrelevant content. As shown in Tables~\ref{tab:gedit_bench}
and~\ref{tab:imgedit}, \ours-6B performs competitively against leading
open-source editors, including several models with substantially more
parameters. With API-based PE, it further achieves the best open-source
result on GEdit-Bench and a leading result on ImgEdit-Bench.
REDEdit-Bench covers a broader range of real-world operations, including
composition, text manipulation, portrait editing, and viewpoint changes.
As reported in Table~\ref{tab:rededit}, \ours-6B with API-based PE achieves
the best overall result among the evaluated open-source models, demonstrating
its ability to execute diverse and complex editing instructions.

\begin{table*}[t]
    \centering
    \caption{
    Results on Qwen-Image-Bench under standard and PE-enhanced prompting. Best and second-best results among open-source models are shown in \textbf{bold} and \underline{underlined}, respectively.
    }
    \label{tab:eval_dimension}
    
    \small
    \renewcommand{\arraystretch}{1.2}
    \begin{tabular}{l c c c c c c}
        \toprule
        \multirow{2}{*}{\textbf{Model}} 
        & \multicolumn{5}{c}{\textbf{Evaluation Dimension}} 
        & \multirow{2}{*}{\textbf{Overall$\uparrow$}} \\
        
        \cmidrule(lr){2-6}
        
        & \textit{Quality} 
        & \textit{Aesthetics} 
        & \textit{Alignment} 
        & \makecell{\textit{Real-world} \\ \textit{Fidelity}} 
        & \makecell{\textit{Creative} \\ \textit{Generation}} 
        & \\
        \midrule
        
        \rowcolor{groupgray}[0pt][0pt]
        \multicolumn{7}{@{}l@{}}{
            \hspace{0.5em}\textit{\textbf{Proprietary / Closed-source Models}}
        } \\

        GPT Image 2
        & 58.65 & 67.53 & 65.85 & 57.38 & 75.23 & 64.69 \\
        
        Nano Banana 2.0
        & 54.77 & 61.08 & 62.40 & 54.28 & 67.05 & 59.82 \\
        
        GPT Image 1.5
        & 55.14 & 60.88 & 61.72 & 53.95 & 66.35 & 59.65 \\
        
        Nano Banana Pro
        & 55.67 & 60.26 & 61.25 & 54.07 & 66.23 & 59.45 \\
        
        Qwen Image 2.0 Pro
        & 54.39 & 58.67 & 59.28 & 51.83 & 64.94 & 57.84 \\
        
        Seedream 5.0
        & 52.55 & 58.40 & 58.90 & 51.92 & 65.29 & 57.22 \\
        
        Seedream 4.5
        & 54.41 & 58.72 & 57.31 & 51.69 & 60.64 & 56.78 \\
        
        Seedream 4.0
        & 54.01 & 58.81 & 56.64 & 51.05 & 58.15 & 56.21 \\
        
        FLUX 2 Max
        & 53.64 & 56.85 & 57.35 & 49.35 & 56.50 & 55.33 \\
        
        FLUX 2 Pro
        & 52.30 & 56.94 & 57.01 & 47.29 & 56.18 & 54.57 \\
        
        GPT Image 1
        & 52.34 & 55.09 & 56.28 & 48.14 & 55.78 & 54.07 \\
        
        Imagen 4.0 Ultra
        & 50.90 & 54.25 & 54.02 & 45.59 & 51.14 & 51.99 \\
        
        Imagen 4.0
        & 50.16 & 52.68 & 51.64 & 44.84 & 47.94 & 50.29 \\
        
        Kling Image 2.1
        & 49.11 & 50.15 & 49.18 & 44.74 & 44.67 & 48.26 \\
        
        \midrule
        
        \rowcolor{groupgray}[0pt][0pt] 
        \multicolumn{7}{@{}l@{}}{
            \hspace{0.5em}\textit{\textbf{Open-source Models}}
        } \\
        
        Qwen Image 2512
        & 51.76
        & 54.74
        & 52.72
        & 47.00
        & 50.19
        & 52.06 \\
        
        Boogu-Image-0.1-Base
        & 51.24
        & 53.50
        & 53.54
        & 46.11
        & 48.91
        & 51.53 \\
        
        HunyuanImage 3.0
        & 50.35
        & 53.57
        & 52.00
        & 44.31
        & 49.12
        & 50.81 \\
        
        Qwen Image
        & 48.44
        & 52.25
        & 50.72
        & 43.16
        & 47.30
        & 49.23 \\
        
        GLM Image
        & 49.26
        & 50.64
        & 47.90
        & 44.69
        & 45.23
        & 48.19 \\

        \ours-3B
        & 51.70
        & 50.96
        & 49.10
        & 44.12
        & 46.57
        & 49.31
        \\
        
        \ours-3B w/ PE
        & 53.15
        & 57.86
        & 55.40
        & 49.28
        & 60.76
        & 55.37
        \\

        \ours-3B w/ PE (API)
        & 52.67
        & \underline{58.03}
        & \underline{59.16}
        & 50.00
        & 61.24
        & \underline{56.44}
        \\
        
        \rowcolor{oursred}
        \textbf{\ours-6B}
        & 51.50
        & 53.01
        & 51.92
        & 45.45
        & 49.65
        & 51.10
        \\
        
        \rowcolor{oursred}
        \textbf{\ours-6B w/ PE}
        & \textbf{53.70}
        & 57.91
        & 56.72
        & \underline{50.69}
        & \underline{63.27}
        & 56.33
        \\

        \rowcolor{oursred}
        \textbf{\ours-6B w/ PE (API)}
        & \underline{53.27}
        & \textbf{60.24}
        & \textbf{60.48}
        & \textbf{51.32}
        & \textbf{64.76}
        & \textbf{58.13}
        \\
        
        \bottomrule
    \end{tabular}
\end{table*}
\paragraph{CPI-General and CPI-Practical.}
CPI-General evaluates comprehensive single- and multi-image editing
capabilities, while CPI-Practical focuses on common real-world applications.
As shown in Tables~\ref{tab:pi_general_single},
\ref{tab:pi_general_multi}, and~\ref{tab:CPI-life-benchmark}, \ours achieves
strong performance across both fundamental editing operations and practical
application scenarios. PE consistently improves both the 3B and 6B models,
with particularly clear gains on multi-reference, compositional, and
application-oriented requests. With API-based PE, \ours-6B achieves the best
CPI-General result among the evaluated open-source models, while
\ours-3B obtains the strongest result on CPI-Practical.

\paragraph{Reasoning-intensive editing.}
CPI-Intelligent evaluates editing tasks that require domain knowledge,
intent inference, and visual layout planning. As shown in
Table~\ref{tab:i2i_reasoning}, direct prompting remains challenging for
existing open-source models. PE substantially improves \ours-3B from
2.02 to 4.10 and \ours-6B from 2.26 to 4.23. With API-based PE,
\ours-6B further reaches 4.24, achieving the best overall result among the
evaluated open-source models. A detailed analysis of PE is provided in the
subsequent subsection.

\paragraph{Compact model.}
\ours-3B incurs nearly no loss after compression and consistently
outperforms the similarly sized FLUX.2-klein-4B on GEdit, ImgEdit, and
REDEdit. With API-based PE, it achieves the second-best open-source overall
score across the five editing benchmarks in
Table~\ref{tab:editing_capability}, closely approaching the 6B variant.
These results demonstrate that the editing capabilities of \ours can be
effectively transferred to a compact model suitable for resource-constrained
deployment.

\subsection{Text-to-Image Evaluation}

\paragraph{Evaluation setup.}
We evaluate the text-to-image generation capability of \ours on
Qwen-Image-Bench\cite{li2026qwen} and our internal evaluation benchmark Pi-ExpertVerse. Qwen-Image-Bench
evaluates production-oriented generation across five dimensions: visual
quality, aesthetics, text--image alignment, real-world fidelity, and creative
generation. Pi-ExpertVerse-T2I complements it by focusing on
knowledge-intensive visual reasoning across diverse expert domains.
\paragraph{Qwen-Image-Bench.}
Qwen-Image-Bench is a creator-centric benchmark designed around real-world professional workflows. In addition to conventional visual quality,
aesthetics, and alignment, it evaluates real-world fidelity and creative
generation, providing a comprehensive assessment of practical T2I
capabilities. As shown in Table~\ref{tab:eval_dimension}, \ours-6B performs competitively under direct prompting. With PE, it achieves the best open-source result overall and leads across all five evaluation dimensions.
Notably, \ours-3B with PE ranks second among the evaluated open-source
models, further demonstrating the effectiveness of our compact architecture.
The PE-enhanced \ours-6B is also competitive with several proprietary
systems.
\paragraph{Reasoning-intensive generation.}
Pi-ExpertVerse-T2I contains 1,000 reasoning-intensive prompts spanning ten
expert domains and 89 fine-grained sub-disciplines. As shown in
Table~\ref{tab:t2i_reasoning}, \ours-6B with PE achieves the best
open-source overall score of 4.63 and leads in nine of the ten expert domains.
\ours-3B with PE also performs strongly, obtaining an overall score of
4.49. A detailed analysis of the effect of PE is provided in the subsequent
subsection.

\begin{table*}[t]
    \centering
        \vspace{6mm}
    \caption{
    Results on Pi-ExpertVerse-T2I under standard and PE-enhanced prompting. Best and second-best results among open-source models are shown in \textbf{bold} and \underline{underlined}, respectively.
    }
    \label{tab:t2i_reasoning}

    \begin{threeparttable}
    \footnotesize
    \setlength{\tabcolsep}{6pt}
    \renewcommand{\arraystretch}{1.2}
    \begin{tabular}{l| c c c c c c c c c c c}
        \toprule
        \textbf{Model}
        & \textbf{IE}
        & \textbf{GH}
        & \textbf{FN}
        & \textbf{DI}
        & \textbf{FB}
        & \textbf{LC}
        & \textbf{SE}
        & \textbf{DP}
        & \textbf{AS}
        & \textbf{NE}
        & \textbf{Overall$\uparrow$} \\
        \midrule
        \rowcolor{groupgray}[0pt][0pt]
        \multicolumn{12}{@{}l@{}}{
            \hspace{0.5em}\textit{\textbf{Proprietary / Closed-source Models}}
        } \\

        GPT-Image-2~\cite{gptimage2_model_card}
        & 4.98 & 4.84 & 4.99 & 4.91 & 5.00 & 4.94 & 4.86 & 5.00 & 4.98 & 4.95 & 4.94 \\
        
        Nano Banana Pro~\cite{google2025nanobanana}
        & 4.86 & 4.62 & 4.79 & 4.74 & 4.81 & 4.83 & 4.65 & 4.81 & 4.92 & 4.88 & 4.78 \\
        
        Seedream5 Pro~\cite{seedream2025seedream}
        & 4.93 & 4.72 & 4.77 & 4.92 & 4.87 & 4.87 & 4.87 & 4.91 & 4.97 & 4.85 & 4.87 \\
        
        Seedream4.5~\cite{seedream2025seedream}
        & 4.38 & 3.71 & 4.22 & 4.01 & 4.14 & 4.05 & 3.78 & 4.40 & 4.42 & 4.10 & 4.10 \\

        Qwen Image 2.0 Pro\cite{zhao2026qwen}
        & 4.61 & 4.21 & 4.43 & 4.32 & 4.20 & 4.25 & 4.37 & 4.51 & 4.61 & 4.46 & 4.39 \\
        
        \midrule
        \rowcolor{groupgray}[0pt][0pt]
        \multicolumn{12}{@{}l@{}}{
            \hspace{0.5em}\textit{\textbf{Open-source Models}}
        } \\

        \multicolumn{12}{@{}l@{}}{
            \hspace{1.5em}\textcolor{gray}{\textit{Standard Evaluation (w/o PE)}}
        } \\
        FLUX.2-klein-4B~\cite{blackforest2025flux2klein} & 3.02 & 2.57 & 2.75 & 1.76 & 2.82 & 1.92 & 1.72 & 2.87 & 2.90 & 2.12 & 2.42 \\
        FLUX.2-klein-9B~\cite{blackforest2025flux2klein} & 3.34 & 2.91 & 2.93 & 1.89 & 2.98 & 2.15 & 1.87 & 3.08 & 3.24 & 2.39 & 2.65 \\
        Qwen-Image-2512~\cite{wu2025qwen} & 3.52 & 2.84 & 3.14 & 2.14 & 2.92 & 2.25 & 1.86 & 3.20 & 3.10 & 2.62 & 2.73 \\
        Boogu-Image-0.1-Base~\cite{chen2026booguimage01} & 3.68 & 3.14 & 3.36 & 2.07 & 3.03 & 2.26 & 1.86 & 3.31 & 3.47 & 2.64 & 2.84 \\
        \ours-3B & 3.01 & 2.56 & 2.85 & 1.72 & 2.37 & 1.93 & 1.62 & 2.73 & 2.79 & 2.41 & 2.37 \\
        \rowcolor{oursred}
        \textbf{\ours-6B} & 3.05 & 2.75 & 2.97 & 1.78 & 2.74 & 1.94 & 1.82 & 2.79 & 3.02 & 2.43 & 2.49 \\

        \multicolumn{12}{@{}l@{}}{
            \hspace{1.5em}\textcolor{gray}{\textit{With Prompt Enhancer (w/ PE)}}
        } \\
        FLUX.2-klein-4B~\cite{blackforest2025flux2klein}
        & 3.91 & 3.28 & 3.54 & 2.37 & 3.50 & 2.73 & 2.48 & 3.80 & 3.75 & 2.83 & 3.19 \\
        
        FLUX.2-klein-9B~\cite{blackforest2025flux2klein}
        & 3.96 & 3.51 & 3.69 & 2.47 & 3.57 & 2.84 & 2.59 & 3.91 & 4.07 & 3.12 & 3.34 \\
        
        Qwen-Image-2512~\cite{wu2025qwen}
        & 4.45
        & 3.88
        & 4.24
        & 3.44
        & 4.22
        & 3.73
        & 3.54
        & 4.48
        & 4.74
        & 4.14
        & 4.04 \\
        
        Boogu-Image-0.1-Base~\cite{chen2026booguimage01}
        & 4.27
        & 3.44
        & 3.89
        & 4.14
        & 3.90
        & 4.11
        & 3.90
        & 4.33
        & 4.35
        & 3.84
        & 4.01 \\
        
        \ours-3B
        & 4.72
        & 4.17
        & 4.49
        & 4.40
        & 4.72
        & 4.50
        & 4.27
        & 4.78
        & 4.88
        & 4.64
        & 4.53 \\

        \ours-3B (API)
        & \underline{4.92}
        & \underline{4.57}
        & \underline{4.73}
        & \underline{4.48}
        & \underline{4.83}
        & \underline{4.81}
        & \underline{4.62}
        & \underline{4.92}
        & \underline{4.96}
        & 4.83
        & \underline{4.75}\\
        
        \rowcolor{oursred}
        \textbf{\ours-6B}
        & 4.81
        & 4.15
        & 4.70
        & 4.46
        & 4.71
        & 4.56
        & 4.59
        & 4.83
        & 4.92
        & \underline{4.86}
        & 4.63 \\

        \rowcolor{oursred}
        \textbf{\ours-6B (API)}
        & \textbf{4.98}
        & \textbf{4.68}
        & \textbf{4.87}
        & \textbf{4.72}
        & \textbf{4.95}
        & \textbf{4.88}
        & \textbf{4.68}
        & \textbf{4.97}
        & \textbf{4.98}
        & \textbf{4.96}
        & \textbf{4.85} \\
        
        \bottomrule
    \end{tabular}

    \begin{tablenotes}[flushleft]
        \footnotesize
        \item[*] Abbreviations: IE = IP \& Entertainment, GH = Geography \& History, FN = Film \& Narrative, DI = Data \& Information, FB = Fashion \& Beauty, LC = Life \& Consumption, SE = Science \& Education, DP = Design \& Product, AS = Architecture \& Space, NE = Nature \& Ecology.
    \end{tablenotes}
    \end{threeparttable}
\end{table*}

\subsection{Effects of Reinforcement Learning and Few-Step Distillation}
We further examine how reinforcement learning and few-step distillation affect the editing capability of \ours. As shown in Table~\ref{tab:editing_posttraining}, RL consistently improves the 6B base model across all evaluated benchmarks, increasing the overall score from 3.98 to 4.16. The gains are particularly pronounced on REDEdit ($3.98\!\rightarrow\!4.34$) and PiPractical ($3.69\!\rightarrow\!3.92$), suggesting that reward-based post-training effectively improves complex instruction following and practical editing quality beyond supervised training. Few-step distillation achieves even stronger overall performance with substantially fewer sampling steps. The resulting \ours-6B-Turbo achieves an overall score of 4.20, slightly exceeding its multi-step RL teacher at 4.16. It closely matches the teacher on GEdit, ImgEdit, and REDEdit, while further improving PiGeneral from 3.96 to 4.08 and PiPractical from 3.92 to 4.01. These results show that our distillation pipeline does not merely recover the base model after sampling compression: it effectively retains the gains introduced by RL and produces a few-step model with comparable or stronger aggregate editing performance.
\begin{table*}[t]
    \centering
    \vspace{6mm}
    \caption{
        Effects of reinforcement learning and few-step distillation on the image editing performance of \ours-6B. Best and second-best results are shown in \textbf{bold} and \underline{underlined}, respectively.
    }
    \label{tab:editing_posttraining}

    \begin{threeparttable}
    \small
    \setlength{\tabcolsep}{8pt}
    \renewcommand{\arraystretch}{1.2}

    \begin{tabular}{l|cccccc}
        \toprule
        \textbf{Model}
        & \textbf{GEdit}
        & \textbf{ImgEdit}
        & \textbf{REDEdit}
        & \textbf{PiGeneral}
        & \textbf{PiPractical}
        & \textbf{Overall$\uparrow$}$^{\ddagger}$ \\
        \midrule

        \ours-6B-Base
        & 7.97
        & 4.41
        & 3.98
        & 3.88
        & 3.69
        & 3.98 \\

        \ours-6B-RL
        & \underline{8.03}
        & \textbf{4.56}
        & \textbf{4.34}
        & \underline{3.96}
        & \underline{3.92}
        & \underline{4.16} \\

        \ours-6B-Turbo
        & \textbf{8.05}
        & \underline{4.55}
        & \underline{4.34}
        & \textbf{4.08}
        & \textbf{4.01}
        & \textbf{4.20} \\

        \bottomrule
    \end{tabular}

    \end{threeparttable}
\end{table*}

\subsection{Prompt Enhancer Evaluation}

\paragraph{Overall effectiveness.}
We further evaluate the effectiveness of the Prompt Enhancer (PE) across
image generation and editing tasks. As shown in
Table~\ref{tab:editing_capability}, PE consistently improves the overall
editing performance of both \ours variants, increasing the scores of the
3B and 6B models from 4.05 to 4.21 and from 4.12 to 4.29, respectively.
A similar trend is observed on Qwen-Image-Bench
(Table~\ref{tab:eval_dimension}), where PE improves the overall T2I scores
from 48.26 to 53.17 for \ours-3B and from 51.10 to 56.33 for
\ours-6B. These consistent gains demonstrate that PE benefits both
generation and editing across different model scales.

\paragraph{Reasoning-intensive tasks.}
The benefits of PE are particularly pronounced on tasks requiring domain
knowledge, intent inference, and visual layout planning. On
CPI-Intelligent, PE improves \ours-3B from 1.99 to 3.13 and
\ours-6B from 2.19 to 4.21. On Pi-ExpertVerse-T2I, the corresponding
scores increase from 2.06 to 4.49 and from 2.49 to 4.63, respectively.
Moreover, applying PE to other open-source models also produces consistent
improvements on both Pi-ExpertVerse splits, indicating that instruction
interpretation and planning are general bottlenecks rather than limitations
specific to the \ours backbone.
\paragraph{Discussion.}
The larger improvements on Pi-ExpertVerse than on conventional benchmarks
support our decoupled design. For straightforward requests, the underlying
DiT already provides strong visual execution, leaving relatively limited room
for prompt refinement. For ambiguous, compositional, or knowledge-intensive
requests, PE resolves user intent, supplements relevant knowledge, and
constructs an executable visual plan before generation. The DiT can therefore
focus on faithful pixel-level rendering rather than simultaneously performing
high-level reasoning and visual synthesis. This gap suggests that complex generation is increasingly limited by high-level intent interpretation and visual planning, rather than rendering capability alone.

\section{Conclusion}
We present Swift-Image, a compact unified model for text-to-image generation and single- and multi-image editing. Through systematic training engineering across architecture, capability-oriented progressive training, reinforcement learning, distillation, and compression, Swift-Image achieves strong open-source performance with only 6B parameters and 243K GPU training hours. A compressed 3B variant retains competitive capabilities, while few-step distillation further reduces inference cost without sacrificing—and in aggregate editing, even improving—performance. Beyond the resulting models, our study highlights several practical principles for compact unified visual generators, including evolving the training distribution with model capability and specializing conflicting objectives before consolidating them into a unified policy. These results suggest that careful coordination across each component can substantially extend the performance frontier of compact visual models. We hope these findings provide a useful reference for building efficient and capable image generation systems.


\newpage

\bibliography{references}

\clearpage
\appendix
\section{Appendix}
\label{sec:appendix}

\subsection{Practical Insights and Takeaways}
\label{sec:practical_insights}

Beyond the specific architectural choices and training recipes presented in the main paper, the development of \ours revealed several general principles that we believe may be useful for building future unified image generation and editing models. 
Rather than treating data construction, architecture design, pre-training, post-training, and deployment optimization as isolated components, our experience suggests that these components should be jointly organized around the \emph{dependency structure of model capabilities}. 
Below, we summarize the main lessons obtained throughout the development process.

\paragraph{Data: Treat the data distribution as a capability curriculum rather than a fixed dataset.}
Different generative capabilities do not emerge simultaneously. Broad visual-semantic alignment and world knowledge are prerequisites for learning fine-grained structure, reference association, and controlled editing. Accordingly, we find it more effective to evolve the data distribution together with the model: early training emphasizes semantic coverage and long-tail concepts, while later stages progressively increase structural complexity, visual fidelity, and task specificity.
This also suggests that data should be organized around transferable capabilities rather than exhaustively enumerating task combinations. For example, text rendering, visual concepts, and stylistic representations learned from T2I data can transfer naturally to editing, reducing the need to independently reconstruct the same visual vocabulary for every editing task.
More generally, the optimal training distribution should be viewed as a function of the model's current capability state rather than as a fixed property of a dataset.

\paragraph{Architecture: Optimize for generative learnability, efficiency, and stability rather than isolated component strength.}
A component that performs well on a standalone benchmark is not necessarily the best component for a generative system.
For example, we observe that a stronger multimodal understanding model does not automatically provide better conditional representations for DiT training. Representations that are highly optimized for query-oriented reasoning may converge more slowly when used as dense generative conditions.
Similarly, architectural choices should be evaluated under a fixed computation or parameter budget. Sharing timestep-modulation parameters slightly reduces the capacity of the modulation branch, but frees a substantial parameter budget that can instead be allocated to the core Attention and MLP layers. In practice, such global reallocation can be more beneficial than locally maximizing the capacity of every component.
Our experience therefore suggests that architectural evaluation should jointly consider downstream convergence, parameter efficiency, operator efficiency, and large-scale training stability.

\paragraph{Pre-training: Coverage should precede preference optimization.}
The primary role of foundational pre-training is to establish a sufficiently broad support of the visual world: semantic concepts, entities, styles, structures, and text-image correspondences.
At this stage, overly aggressive quality filtering can be counterproductive, since visually imperfect images may still contain valuable semantic supervision. When such imperfections are accurately described, they can even become recognizable visual concepts that the model learns to avoid reproducing.
We therefore prioritize semantic coverage and distributional diversity before aggressively concentrating probability mass on aesthetically preferred outputs.
From this perspective, pre-training expands the support of the learned distribution, while subsequent stages reshape probability mass within that support.

\paragraph{Pre-training curriculum: Increase spatial computation together with supervision complexity.}
Progressive resolution is most effective when it is coupled with increasing information density.
Low-resolution training is sufficient for learning broad semantic correspondences and global structure, whereas dense text, complex layouts, and fine-grained spatial relationships become substantially more informative only at higher resolutions.
Thus, increasing resolution should not merely repeat the same supervision at a higher computational cost; it should coincide with the introduction of visual content whose structure actually requires the additional spatial bandwidth.
Similarly, unified generation and editing do not need to be jointly optimized from the beginning. In our training pipeline, editing is introduced only after stable T2I capability has been established, allowing editing to reuse an already learned visual vocabulary.

\paragraph{Continual pre-training: View CT as distribution annealing from breadth to quality.}
Continual pre-training serves a distinct role between large-scale pre-training and supervised fine-tuning.
Its goal is not simply to continue optimization for more steps, but to gradually move the model from a broad and noisy data distribution toward a cleaner, higher-fidelity distribution while preserving world knowledge and task coverage.
An important practical observation is that quality filtering itself introduces distribution shift: professional and high-fidelity data sources tend to have narrower semantic and stylistic distributions than web-scale corpora.
We therefore find it important to couple quality refinement with explicit semantic rebalancing.
In general, every substantial increase in data quality should be accompanied by a mechanism that preserves coverage and prevents long-tail knowledge from being unintentionally removed.

\paragraph{Supervised fine-tuning: Use SFT to select a high-quality sub-manifold rather than relearn the world.}
After sufficiently broad pre-training and continual pre-training, SFT is most effective when using a relatively small but highly curated dataset.
Its purpose is to guide the model toward a desirable high-quality sub-manifold characterized by stronger visual fidelity, instruction following, and human preference alignment, rather than to introduce large amounts of new world knowledge.
Moreover, the most valuable SFT samples are not necessarily the globally highest-quality examples, but those that directly address the model's remaining capability gaps.
This motivates our evaluation-driven data loop: recurring failures are identified by capability-aware evaluation, converted into retrieval or synthesis targets, and assigned higher sampling weights until the corresponding gaps are resolved.
Under this view, evaluation becomes an active control signal for training rather than merely a terminal measurement.

\paragraph{Reinforcement learning: Parallel RL and multi-expert consolidation.}
Image-generation RL must jointly optimize heterogeneous objectives spanning semantic alignment, perceptual quality, and task-specific fidelity. A single mixed-task policy often suffers from gradient interference and seesaw trade-offs, while limited sample efficiency forces under-optimized domains toward longer training horizons, where diffusion RL becomes increasingly unstable. Our key insight is to decouple specialization from unification. We retain a mixed-task policy as a generalist anchor and train task-coherent experts in parallel with domain-routed reward combinations, allowing each policy to approach its domain-specific performance ceiling under a coherent objective. We then consolidate their complementary gains into a shared student through multi-teacher on-policy distillation, preserving generalist coverage while absorbing specialist strengths. Therefore, \emph{specialized capabilities need not be learned and unified within the same optimization trajectory}: parallel RL can optimize them independently, and multi-expert consolidation can integrate them afterward.


\paragraph{Distillation: What matters is which states you distill on, and not mixing conflicting tasks.}
Since a few-step model only ever runs on its own trajectory, we build distillation states from student rollouts. The key question is then \emph{which} states to use. The rollout does not need to reproduce the inference schedule---it only needs to cover it, so we intentionally randomize the training-time transitions, which lets one student serve multiple step budgets. Supervision should not be spread evenly over noise levels either: high-noise states mainly fix global structure, low-noise states fix details and texture. Because a compressed trajectory leaves no later steps to repair earlier mistakes, errors made at high noise are essentially irreversible, so instead of treating all states uniformly we put supervision where the student is still weak, and revisit this allocation as the student improves. Finally, T2I and editing interfere during distillation: T2I mainly needs diversity and global structure, whereas editing needs reference fidelity and consistency in unedited regions, and distilling both together gives a student that is worse at both. We therefore distill them separately, let each use its own state distribution, and consolidate afterwards, as in RL.

\paragraph{Prompt Enhancer: Treat prompt enhancement as compilation into the generator's native conditional language.}
A central lesson from our system is that a Prompt Enhancer should not be regarded merely as a verbose prompt rewriter.
During generator training, the Captioner defines the language distribution observed by the DiT, including its vocabulary, granularity, spatial descriptions, image-reference conventions, and layout representations.
At inference time, raw user requests may follow a substantially different distribution.
The role of the Prompt Enhancer is therefore to translate heterogeneous user intent into a generator-aligned conditional representation that resembles the supervision language seen during training.
In this sense, prompt enhancement is fundamentally a \emph{condition-distribution alignment} problem.

For complex requests, we further find it beneficial to decouple semantic reasoning from pixel-level execution.
The Prompt Enhancer interprets ambiguous instructions, introduces necessary knowledge, resolves reference roles, and constructs an explicit visual plan, while the DiT focuses on faithfully rendering the resulting condition.
Reasoning is activated adaptively: straightforward requests are directly compiled, whereas knowledge-intensive or layout-sensitive tasks use additional reasoning.
Finally, prompt quality should not be optimized solely through textual judgment. A prompt that appears semantically complete may still be difficult for a particular generator to execute.
Generator-aware reinforcement learning therefore evaluates both the rewritten condition itself and the image actually produced by the frozen generator.

\paragraph{A unified perspective.}
These observations can be summarized by five broader principles:

\begin{enumerate}
    \item \textbf{Capability Dependency.}
    Model capabilities have a natural dependency structure; data composition, resolution, and optimization objectives should evolve accordingly.

    \item \textbf{Distribution Evolution.}
    Training progresses from broad and noisy coverage toward structured, balanced, clean, and preference-aligned distributions rather than repeatedly optimizing on a fixed corpus.

    \item \textbf{Specialize--Consolidate.}
    When heterogeneous objectives strongly interfere, first optimize coherent specialist policies and subsequently consolidate their strengths into a unified model.

    \item \textbf{Reason--Render Decoupling.}
    Semantic interpretation, knowledge reasoning, and visual planning can be handled separately from pixel-level synthesis, allowing each component to specialize in the form of computation it performs best.

    \item \textbf{Evaluation as Control.}
    Evaluation should continuously identify capability gaps and modify the training distribution, turning benchmarking from a terminal measurement into a closed-loop optimization signal.
\end{enumerate}

Overall, our experience suggests that the development of a strong unified visual foundation model is less about identifying a single dominant architectural or optimization trick, and more about coordinating the evolution of \emph{data, capability, computation, and optimization} throughout the entire training lifecycle.

\subsection{Architecture Choices and Ablations}
\label{app:architecture}

\paragraph{Architectural Design Principles}
To determine the optimal architectural configuration, we conducted exhaustive architectural ablation studies. Drawing inspiration from the design philosophy of Krea2\citep{krea-2-2026}, our selection of ablation components was comprehensively evaluated against the following three core objectives:
\begin{itemize}[leftmargin=*, topsep=0pt]
    \item \textbf{Model Performance}: Under an identical training TFLOPs budget, does the component accelerate model convergence? Does it elevate the model's ultimate performance ceiling over extended training cycles? Furthermore, how well does the design generalize across diverse task categories?
    \item \textbf{Model Efficiency}: Without compromising performance, can the component significantly reduce computational FLOPs or enhancing the Model FLOPs Utilization (MFU) during GPU training?
    \item \textbf{Training Stability}: Does the component contribute to smoother curves for both Training Loss and Gradient Norm?
\end{itemize}

Our ablation designs extensively reference the architectural choices of leading generative models\cite{blackforest2025flux2klein,wu2025qwen,cai2025z,zhuo2024lumina,krea-2-2026} and LLMs\cite{yang2025qwen3} in the open-source community. The established frameworks of these large-scale models provided crucial guidance for this work. The baseline settings, ablation candidates, and the final selections for each core component are summarized in Table~\ref{tab:ablation_summary}.
\begin{table}[t]
    \centering
    \caption{Summary of architectural component ablations and final selections.}
    \label{tab:ablation_summary}

    \small
    \setlength{\tabcolsep}{3.5pt}
    \renewcommand{\arraystretch}{1.2}

    \begin{tabularx}{\linewidth}{
        @{}
        P{0.16\linewidth}
        P{0.22\linewidth}
        Y
        P{0.25\linewidth}
        @{}
    }
        \toprule
        \textbf{Component}
        & \textbf{Baseline}
        & \textbf{Ablated variants}
        & \textbf{Final choice} \\
        \midrule

        \textbf{Attention}
        & Multi-head attention\citep{vaswani2017attention}
        & Conditional feature-reuse attention\citep{tan2026ominicontrol2}
        & Multi-head attention \\

        \textbf{MLP}
        & GeLU MLP
        & SwiGLU\citep{shazeer2020glu}
        & SwiGLU \\

        \textbf{Residual}
        & Standard residual
        & Kimi Attention Residual\citep{team2026attention}
        & Standard residual \\

        \textbf{Text encoder}
        & Qwen2.5-VL-7B\citep{bai2025qwen2}
        & Qwen3-VL-4B/8B\citep{yang2025qwen3};\newline
          Qwen3.5-4B\citep{qwen3.5}
        & Qwen3-VL-8B \\

        \textbf{Modulation}
        & Per-block MLP
        & Block-shared modulation
         \citep{blackforest2025flux2klein,rock2025chroma}
        & Block-shared modulation \\

        \textbf{Auto-encoder}
        & FLUX AE\citep{flux2024}
        & Qwen-Image VAE\citep{wu2025qwen};
          FLUX.2 AE\citep{blackforest2025flux2klein};
          Internal AE
        & FLUX.2 AE \\

        \textbf{Block design}
        & Single-stream Transformer
        & Hybrid stream\citep{esser2024scaling};
          Single-stream parallel
         \citep{dehghani2023scaling,blackforest2025flux2klein}
        & Single-stream parallel \\

        \textbf{Norma-lization}
        & LayerNorm\citep{ba2016layer}
        & RMSNorm\citep{zhang2019root};
          Hybrid norm;
          Sandwich norm\citep{zhuo2024lumina}
        & QK: RMSNorm; others: LayerNorm; sandwich norm \\

        \textbf{Position encoding}
        & 3D-RoPE
        & MS-RoPE\citep{wu2025qwen};
          4D-RoPE\citep{blackforest2025flux2klein}
        & 4D-RoPE \\

        \textbf{Prompt encoding}
        & Default tokenization; text preceding; retain VLM image features
        & Character-level tokenization; \newline image preceding; discard VLM image features
        & Character-level tokenization; image preceding; discard VLM image features \\

        \bottomrule
    \end{tabularx}
\end{table}
\paragraph{Text Encoder}
Given the DiT community's overarching shift towards employing MLLMs as text encoders\citep{wu2025qwen, krea-2-2026,song2026joyai}, we conducted ablation tests on four distinct encoders: Qwen2.5 VL-7B\cite{bai2025qwen2}, Qwen3 VL-4B/8B\cite{yang2025qwen3}, and Qwen3.5-4B\cite{qwen3.5}. Empirical results demonstrated that Qwen3 VL-8B delivered the best overall performance. Notably, although Qwen3.5-4B outperformed Qwen3 VL-8B on public benchmarks for multimodal understanding, its actual training convergence was significantly delayed. We hypothesize that because Qwen3.5-4B is inherently designed as a ``Thinking Model,'' its output representations are biased toward a query-oriented logic rather than the dense semantic information essential for generation tasks\cite{ma2024exploring}. Ultimately, we selected Qwen3 VL-8B as our text encoder, which possesses richer multimodal and multilingual representational capabilities. To capture multi-scale semantic information\cite{li2025unifusion}, we extracted and concatenated representations from three uniformly distributed hidden layers within the model, subsequently aligning them to the DiT latent space via an MLP connector.

\paragraph{Autoencoder}
High-quality latent space representations can significantly accelerate the convergence of DiT models\cite{zheng2025diffusion, yu2024representation, wu2026representation,yue2026matters}. We compared the FLUX AE\cite{flux2024}, Qwen Image VAE\cite{wu2025qwen}, FLUX 2 AE\cite{blackforest2025flux2klein}, and an in-house representation encoder PAE\cite{yue2026matters} (as summarized in Table~\ref{tab:ablation_summary}). Evaluating both reconstruction quality and downstream DiT training convergence, FLUX 2 AE achieved the best performance and was thus selected as the final solution. It is worth mentioning that while our internally trained AE exhibited faster convergence in the early stages, it introduced noticeable text and image artifacts into the generated results. We attribute this to inherent reconstructability limitations within this representation encoder, which constrained the model's convergence ceiling. We plan to specifically optimize the reconstructability of the internal AE in future work.

\paragraph{Transformer Block}
A standard DiT block consists of serialized Attention and Multi-Layer Perceptron (MLP) layers. We explored ablations for both:
\begin{itemize}[leftmargin=*, topsep=0pt]
    \item \textbf{Attention Layer}: For image editing tasks, we experimented with the \textit{Conditional Feature Reuse} mechanism proposed by OmiControl2\cite{tan2026ominicontrol2}. This involves caching the KV tokens of the original reference image during the first denoising step and sharing this cache in subsequent steps to reduce computational overhead. However, our experiments indicated that this operation led to performance degradation and slower convergence. Consequently, this design is excluded from the current version, though related optimizations remain under active exploration.
    \item \textbf{MLP Layer}: Following practices in current mainstream DiT frameworks\cite{krea-2-2026,blackforest2025flux2klein}, we replaced the MLP activation function with SwiGLU (visualized within the MLP-In module in Figure~\ref{fig:arch}). Drawing on the FLUX-Kelin\cite{blackforest2025flux2klein} configuration, we set the SwiGLU expansion factor to 3. This modification yielded steady performance improvements.
\end{itemize}

Additionally, regarding the multimodal stream design within the MM-DiT, we compared three architectures:
\begin{enumerate}[leftmargin=*, topsep=0pt]
    \item \textbf{Pure Single-stream}: Tokens from different modalities fully share the QKV projection matrices and MLP parameters.
    \item \textbf{Hybrid Stream}: Tokens from different modalities utilize independent QKV and MLP parameters in the shallow layers, but transition to shared parameters in the deeper layers.
    \item \textbf{Parallel Single-stream}: Abandoning the serialized paradigm of standard Transformers, this design employs a parallel computation strategy for the Attention and MLP modules\cite{dehghani2023scaling,blackforest2025flux2klein} (as detailed in the right panel of Figure~\ref{fig:arch}), the operation can be simplified as:
    \begin{equation*}
        y' = \text{LayerNorm}(x), \quad y = x + \text{MLP}(y') + \text{Attention}(y')
    \end{equation*}
    Through the fusion of linear projection matrices, this design achieves additional computational parallelization. Specifically, the projection matrix multiplications for Query, Key, and Value are fused into a single operator with the first linear transformation of the MLP. Similarly, the projection of the attention output is fused with the second linear transformation of the MLP.
\end{enumerate}
Experiments revealed that, under an identical number of layers, the Hybrid Stream architecture slightly outperformed both the Pure Single-stream and Parallel Single-stream designs, albeit at the cost of additional parameter overhead. Although the Parallel Single-stream architecture incurred an exceedingly small performance penalty compared to its non-parallel counterpart, its operator fusion capability improved inference efficiency by approximately 10\%. Weighing efficiency against architectural simplicity, we ultimately adopted the Parallel Single-stream architecture.

\paragraph{Timestep Conditioning}
In standard MM-DiT models\cite{esser2024scaling}, each Transformer block contains an independent MLP to generate the scale, shift, and gate factors for the current layer. These MLP modules typically account for $\sim$ 20\% to 30\% of the total parameter count. The community consensus is that consuming such a massive parameter budget merely to inject scalar conditions is highly inefficient\cite{rock2025chroma,cai2025z}. Therefore, we substituted the independent MLPs with a \textit{Block-shared Modulation} strategy\cite{blackforest2025flux2klein}. Although removing these parameters directly caused a small drop in performance, this optimization freed up a substantial parameter budget. Under the same total parameter constraint, this allowed us to allocate more parameters to the core Attention and MLP layers, thereby effectively elevating the global model performance.
\paragraph{Positional Encoding}
For Rotary Positional Encoding (RoPE)\cite{su2024roformer}, we conducted multiple comparative experiments. The baseline model utilized 3D-RoPE (encoding Frame, Height, and Width respectively, with the index offset for adjacent reference images set to 1 in multi-image editing tasks). Ablation candidates included MS-RoPE (proposed by Qwen Image\cite{wu2025qwen}) and 4D-RoPE (adopted by FLUX 2\cite{blackforest2025flux2klein}). Final results demonstrated that 4D-RoPE exhibited the most superior resolution extrapolation capabilities. In this configuration, FLUX 2 4D-RoPE spans four dimensions: \texttt{[T, H, W, L]}. Text tokens are exclusively encoded in the Language (L) dimension, with the remaining dimensions set to zero. When encoding adjacent images, a mandatory index offset of 10 units is applied in the Time (T) dimension. This strategy introduces an image index prior and mitigates the over-attention of target image tokens to tokens at identical spatial locations in adjacent images, thereby circumventing ``copy-paste'' artifacts during generation\cite{xia2025dreamomni2}.

\paragraph{Residual \& Normalization}
\begin{itemize}[leftmargin=*, topsep=0pt]
    \item \textbf{Residual Connection}: We retained the standard residual connection. While we experimented with the \textit{Kimi-Attn Residual}\cite{team2026attention} design, it yielded no performance gains in our image generation model and instead caused performance degradation. The exact reasons warrant further investigation, but we preliminarily surmise that the highly unstructured features of image data are incompatible with the Kimi-Attn Residual mechanism\cite{liu2026AttnRes}.
    \item \textbf{Normalization Strategy}: Uniformly replacing LayerNorm with RMSNorm across the entire network led to a slight performance dip. Therefore, we adopted a hybrid strategy (as depicted in Figure~\ref{fig:arch}): utilizing RMSNorm solely for the QK-Norm (\texttt{Rms-Norm} block within Attention), while retaining LayerNorm for the remaining components. Furthermore, we introduced Sandwich-Norm to constrain the signal magnitudes at the inputs and outputs of the Attention modules and Feed-Forward Networks, and applied a Tanh function to the Gate parameters to prevent abnormally high modulation values from destabilizing the residual branches\cite{zhuo2024lumina,cai2025z}. This combination of strategies significantly enhanced training stability, manifesting as smoother Loss curves and Gradient Norms.
\end{itemize}

\paragraph{Optimizer}
Throughout our current training pipeline, AdamW served as the primary optimizer. Concurrently, we explored the application potential of the Muon optimizer\cite{liu2025muon} on the MM-DiT architecture, utilizing its standard implementation in the DeepSpeed framework. We observed that due to Muon's orthogonalization strategy, its parameter update magnitudes are inherently larger on a numerical scale; thus, disabling Gradient Clipping yielded better gains in this scenario. During our exploratory experiments, Muon indeed converged faster than AdamW in the initial training stages, but its ultimate performance over extended timescales fell short of AdamW. Drawing on recent community practices\cite{krea-2-2026}, the optimized parameter set for Muon requires meticulous filtering. Due to time constraints, we have yet to integrate the refined Muon scheme into our most recent formal pre-training round, but this remains a focal point for our next phase of optimization.

\paragraph{Prompt Tokenization and Conditional Encoding}
We implemented optimizations and unifications regarding the processing mechanism of conditional inputs across the following three dimensions:
\begin{itemize}[leftmargin=*, topsep=0pt]
    \item \textbf{Character-level Tokenization}: For ``text to be rendered'' demarcated by quotation marks within the Prompt, we applied a character-level tokenization strategy (while keeping default tokenization for other text)\cite{team2025longcat}. Experiments revealed that this strategy substantially improved the precision of text rendering in generated images.
    \item \textbf{Modality Ordering}: For interleaved image-text editing instructions, prepending image data before the editing text as input to the VLM yielded minor yet stable performance gains. We postulate this is because the model acquires a purer initial image representation alongside a more globally contextualized instruction representation.
    \item \textbf{Image Token Discarding}: The Image Token representations output by the VLM encoder are not forwarded to the downstream DiT network. Ablation results confirmed that discarding these tokens not only incurred no performance loss but also accelerated the overall inference speed (this design is explicitly indicated by the ``drop'' arrow above the VLM in Figure~\ref{fig:arch}).
    \item \textbf{Unified System Prompt}: In previous research, divergent System Prompt templates were typically deployed across different tasks to acquire task-specific representations and avert inter-task conflicts\cite{wu2025qwen,ma2024exploring}. However, our ablation tests demonstrated that utilizing a globally unified template did not compromise performance. Adhering to the principle of architectural simplicity, we applied an identical System Prompt across all supported tasks.
\end{itemize}

\subsection{Progressive Training Details}
\label{app:progressive_training}
\subsubsection{Pre-training}

\begin{figure*}[t]
\centering
  \includegraphics[width=1.0\textwidth]{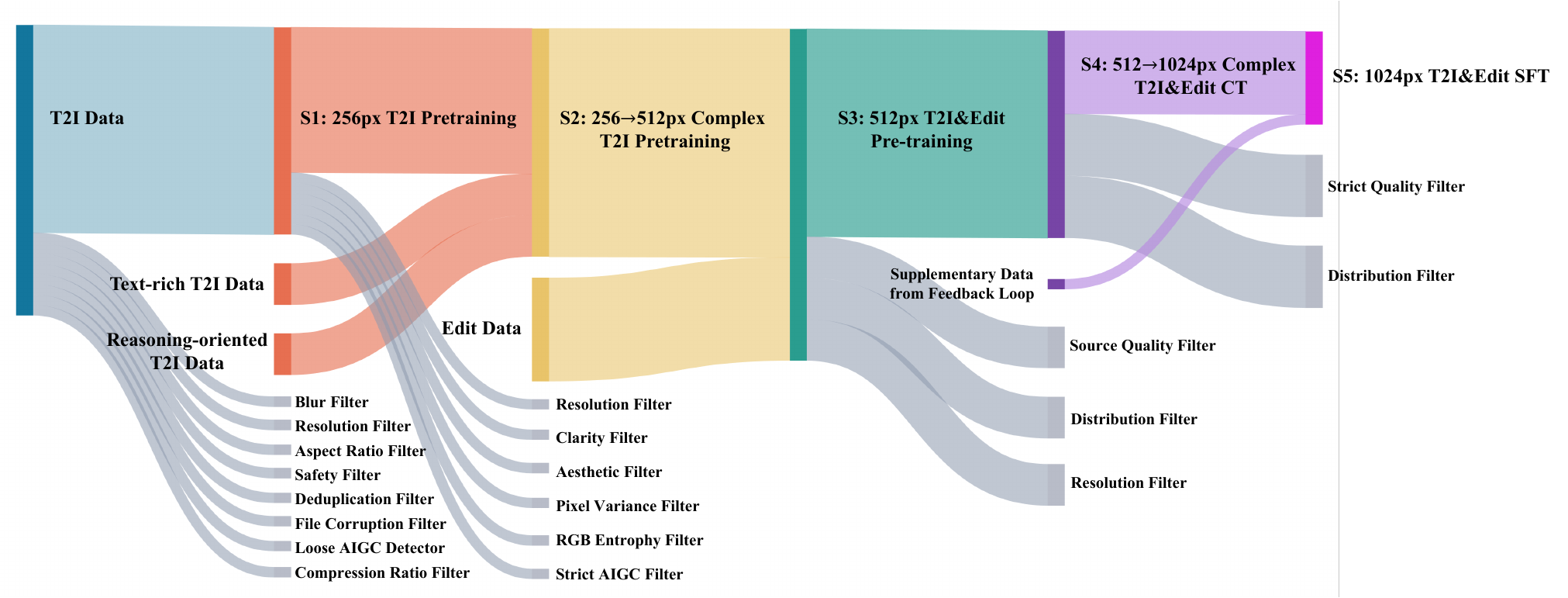}
 
  \caption{Data flow and stage-specific filtering throughout progressive training.}
  \label{fig:training_data_pyramid}
 
\end{figure*}

Foundational pre-training aims to establish broad visual concepts, world knowledge, text-image alignment, and structural modeling. In contrast to later stages that increasingly emphasize visual quality and preference alignment, this phase prioritizes data scale, semantic coverage, and diversity. We train for 500K steps.

\paragraph{Multi-stage Curriculum Learning}
We adopt a three-stage curriculum consisting of low-resolution T2I pre-training, progressive-resolution T2I pre-training, and multi-task joint pre-training. Resolution is first increased from 256px to 512px before introducing greater task complexity, decoupling spatial scaling from multi-task learning. This curriculum gradually transitions the model from basic text-image semantics to fine-grained visual modeling and unified generation-editing capabilities (Fig.~\ref{fig:training_data_pyramid}).

\paragraph{Resolution Bucketing and Task-routed Data Packing}
To efficiently train on diverse aspect ratios and highly heterogeneous prompt lengths, we introduce two framework-level optimizations:
\begin{itemize}[leftmargin=*, topsep=0pt]
    \item \textbf{Aspect-ratio bucketing.}
    Images with similar aspect ratios are assigned to predefined resolution buckets, reducing unnecessary padding or cropping while supporting generation across diverse aspect ratios.

    \item \textbf{Task-routed sequence packing.}
    Prompt lengths vary substantially across tasks: standard T2I and simple editing samples are typically short, whereas dense text-rendering and complex editing samples can exceed \(10{,}000\) tokens. We therefore route tasks according to their sequence-length characteristics. Short-prompt tasks use conventional batching with padding to 512 tokens, while long-prompt tasks are packed along the sequence dimension and processed with the \texttt{flash\_attn\_var\_len\_func} interface of FlashAttention-2. This avoids excessive padding or truncation and improves computation efficiency for extremely long conditions, particularly under our single-stream architecture.
\end{itemize}

\subsubsection{Continual Pre-training}

Continual Pre-training (CT) bridges large-scale noisy pre-training and high-quality Supervised Fine-Tuning (SFT). We train for 200K steps while progressively increasing the resolution from 512px to 1024px, with the goal of improving the visual quality baseline without sacrificing semantic coverage.

\begin{itemize}[leftmargin=*, topsep=0pt]
    \item \textbf{High-quality data curation.}
    We remove lower-quality web-crawled sources and emphasize visually strong or professionally curated domains, improving visual fidelity while retaining broad world knowledge and stylistic diversity.

    \item \textbf{Semantic balancing with VLM-based categorization.}
    To prevent high-quality sources from introducing distribution bias, we use VLMs to construct hierarchical semantic labels and resample the data according to this taxonomy. The resulting distribution improves category balance and coverage of long-tail concepts.

    \item \textbf{Fine-grained editing data balancing.}
    Editing data undergo additional quality filtering and category-level balancing across instruction types, reducing task imbalance and catastrophic forgetting during joint generation-editing training.

    \item \textbf{Progressive resolution scaling.}
    Training resolution is gradually increased from 512px to 1024px together with the shift toward higher-quality data, allowing the model to devote more capacity to high-frequency visual details.
\end{itemize}

\begin{table}[t]
\centering
\caption{Configurations and hyperparameters for progressive training.}
\label{tab:training_configurations}
\small
\resizebox{\columnwidth}{!}{
\begin{tabular}{lccc}
\toprule
\textbf{Configuration} & \textbf{Pre-training} & \textbf{Continual Pre-training} & \textbf{Supervised Fine-tuning} \\
\midrule
\multicolumn{4}{l}{\textbf{Training Process}} \\
Steps (K) & 500 & 200 & 10 \\
Resolution & 256/512 & 512/1024 & 1024 \\
Batch Size (K) & 8 & 8/4 & 4 \\
\midrule
\multicolumn{4}{l}{\textbf{Data Distribution}} \\
Type & T2I/TI2I & T2I/TI2I & T2I/TI2I \\
Ratio & 7/1 & 7/3 & 7/3 \\
\midrule
\multicolumn{4}{l}{\textbf{Hyperparameters}} \\
Optimizer & AdamW & AdamW & AdamW \\
Weight Decay & 0.01 & 0.01 & 0.01 \\
Grad. Norm Clip & 1.0 & 1.0 & 1.0 \\
Uncond. Dropout & 0.1 & 0.1 & 0.1 \\
Learning Rate & $1\times 10^{-4}$ & $2\times 10^{-5}$ & $1\times 10^{-5}$ \\
\bottomrule
\end{tabular}
\vspace{-3mm}}
\end{table}

\subsubsection{Supervised Fine-Tuning}

SFT further improves visual fidelity and instruction following by concentrating training on a small, highly curated data distribution. We train for 10K steps at 1024px, prioritizing data quality over scale.

\begin{itemize}[leftmargin=*, topsep=0pt]
    \item \textbf{Human-AI collaborative curation.}
    We sample only from top-quality sources and apply a two-stage filtering pipeline: large VLMs first remove samples with evident visual defects or text-image misalignment, followed by human review for final quality control.

    \item \textbf{Evaluation-driven data enhancement.}
    Based on evaluation results from preceding stages, we selectively increase data for underperforming tasks and semantic concepts. Global category balancing is retained to avoid domain overfitting and preserve long-tail capabilities.

    \item \textbf{Timestep sampling optimization.}
    Pre-training uses Logit-Normal timestep sampling to emphasize medium-noise regions and facilitate learning of global image structure. During SFT, we switch to Uniform Sampling to provide balanced supervision across timesteps and strengthen fine-grained texture and detail modeling.
\end{itemize}

Pretraining and supervised fine-tuning (SFT) endow the model with strong and generalizable capabilities for text-to-image generation and instruction-based image editing. However, supervised objectives are not sufficient to fully capture human preferences, such as style consistency, compositional coherence, overall aesthetics, and stable prompt fidelity.

Reinforcement learning from human feedback (RLHF) and on-policy distillation have shown strong effectiveness in aligning large language models, and offer a principled approach for multimodal generation as well. Motivated by this, recent work increasingly adopts reinforcement learning as a post-training strategy to further improve generation quality, instruction following, realism, and edit fidelity.

Despite this promise, applying reinforcement learning to multimodal generation introduces several practical challenges. (1) Reward evaluation is inherently multi-objective. High-quality and discriminative reward signals require both well-trained reward models and a systematic reward evaluation pipeline. (2) Existing training paradigms such as Flow-GRPO and DiffusionNFT are often limited in the amount of data and the number of optimization steps that can be effectively covered in a single RL run. Excessively long training may lead to mode collapse, limiting training scalability. (3) Different tasks may exhibit significant gradient conflicts. Consequently, mixed-task RL training often suffers from a seesaw effect across task categories.

To address these issues, we build a multi-dimensional online reward scoring system for real-time reward feedback during training. On top of this system, we propose a two-stage cascaded reinforcement learning pipeline based on DiffusionNFT and multi-teacher on-policy distillation (OPD). We first optimize mixed-task data using the full set of multi-dimensional rewards to obtain a strong general-purpose model. We then train domain-specific teacher models with tailored reward designs and specialized data for each expert domain. Finally, we use multi-teacher OPD to consolidate the strengths of these domain teachers into a single final model.

\subsubsection{Reward System}
\begin{itemize}[leftmargin=*, topsep=0pt]
\item \textbf{T2I-Reward} For text-to-image (T2I) tasks, we develop a rubric-based multi-dimensional reward system consisting of \textit{Text-Image Alignment}, \textit{category-aware aesthetic scoring}, \textit{visual quality assessment}, and \textit{prompt-based style consistency evaluation}. 
Text-Image Alignment measures overall prompt fidelity; aesthetic scoring captures human preference across different image types, such as design-centric and photorealistic images; visual quality assessment evaluates structural plausibility, including anatomy, text rendering, and object integrity; and style consistency evaluates whether the generated image faithfully matches the prompt-specified style, including style alignment and photographic realism.

\item \textbf{Edit-Reward} For image editing tasks, our rubric-based reward system primarily emphasizes \textit{instruction following}, \textit{consistency with the reference image}, and \textit{visual quality assessment}. To further enhance the discriminative capability and stability of online reward scoring, we incorporate an auxiliary preference reward model trained with BT loss. In addition, for domain-specific expert teacher training, we specially customize the reward criteria according to the characteristics of different editing domains.

\item \textbf{Supplementary Rewards} We further incorporate supplementary reward signals to improve face-identity consistency, enhance text rendering accuracy, and reduce distorted glyph generation. Specifically, we use an ArcFace-based facial representation model~\citep{deng2019arcface} to assess character identity consistency. Recent advances~\citep{zhang2024ssmgao,zhang2025syntabllavagao,zhang2024darling,zhang2024control,zhang2026ppocrv6} in visual text recognition, editing, generation, and structured understanding provide a strong foundation for reward-based evaluation of text-rendering accuracy. Accordingly, we adopt PP-OCRv6~\citep{zhang2026ppocrv6} as an auxiliary reward model for assessing text-rendering accuracy.

\subsection{Expert Reinforcement Learning and OPD Details}
\label{app:post_training}
\begin{figure*}[t]
    \centering
    \includegraphics[width=1\linewidth]{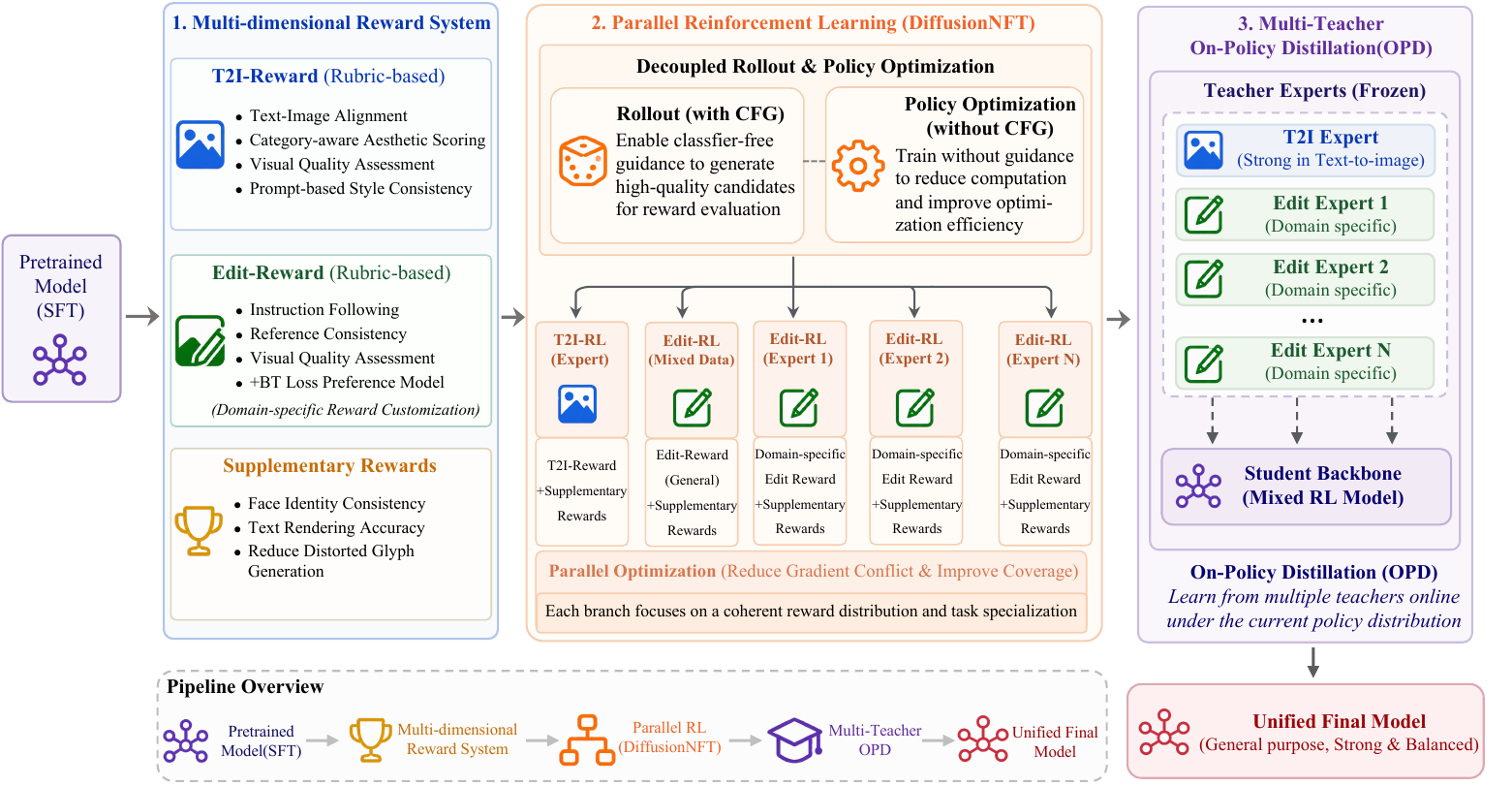}
    \caption{Parallel expert reinforcement learning and multi-teacher on-policy distillation. Multiple policies are optimized with task-specific rewards and then consolidated into one unified model.}
    \label{fig:rl_pipeline}
\end{figure*}

\end{itemize}

\subsubsection{Parallel Reinforcement Learning}
Reinforcement learning for diffusion and flow-based generation is fundamentally constrained by low sample efficiency and limited training stability. In practice, methods such as DiffusionNFT~\cite{zheng2025diffusionnft} and Flow-GRPO~\cite{liu2025flow} often require many optimization steps before yielding consistent gains, while overly long RL training can introduce instability, including training collapse and mode collapse. As a result, a single RL run can only expose the model to a limited amount of effective training data. This limitation becomes more severe under mixed-task training, where heterogeneous objectives induce gradient interference, making it difficult for one policy to achieve strong performance across all tasks.

To address this issue, we perform parallel reinforcement learning based on DiffusionNFT rather than relying on a single mixed-task RL run. Specifically, we train a dedicated T2I-RL model, a mixed-data Edit-RL model, and several expert policies for editing sub-tasks that remain under-optimized in joint training. This parallel design reduces cross-task optimization conflict and allows each branch to focus on a more coherent reward distribution, thereby improving both task specialization and overall coverage.

We further optimize the DiffusionNFT training procedure by decoupling rollout and policy optimization. During rollout, classifier-free guidance is enabled to improve sample quality and produce stronger candidates for reward evaluation. During policy optimization, training is performed without guidance to reduce computational cost and improve optimization efficiency. This design preserves the benefit of guidance-enhanced reward signals while avoiding the full overhead of guided policy training, and implicitly distills the advantages of guided sampling into the final policy.

\subsubsection{Multi-Teacher On Policy Distillation}
In both large language models and generative modeling, Multi-teacher On Policy distillation~\citep{agarwal2024onpolicy,xiao2026mimov2flashtechnicalreport,deepseekai2026deepseekv4,ma2026mopd,zhou2026danceopd,li2026diffusionopd} has increasingly emerged as an effective paradigm for consolidating multiple domain-specialized expert models into a single general-purpose model. After the previous stage of fine-grained domain partitioning and expert RL training, we obtain a collection of RL experts specialized in different tasks and subdomains. Among them, the RL model trained on mixed-task data serves as a strong generalist: it achieves balanced performance across diverse tasks and maintains robust overall capability. However, such balanced optimization also limits its ability to fully reach the performance ceiling on certain subdomains. In contrast, the expert trained specifically for text-to-image generation, as well as experts specialized for under-optimized editing domains, are able to achieve substantially stronger performance within their respective areas.

Motivated by this observation, we adopt a multi-teacher on-policy distillation to integrate these specialized capabilities into a unified model. Specifically, we take the mixed-data RL model as the student backbone due to its strong generalization and balanced multi-task performance, and then distill into it the knowledge of multiple domain experts, including the T2I expert and several experts for different image editing subdomains. In this way, the final model preserves the broad coverage and task balance of the mixed-data RL model, while further absorbing the domain-specific strengths of specialized expert teachers.

\subsection{Compression and Few-Step Distillation Details}
\label{app:compression}

\begin{figure*}[t]
    \centering
    \includegraphics[width=1\linewidth]{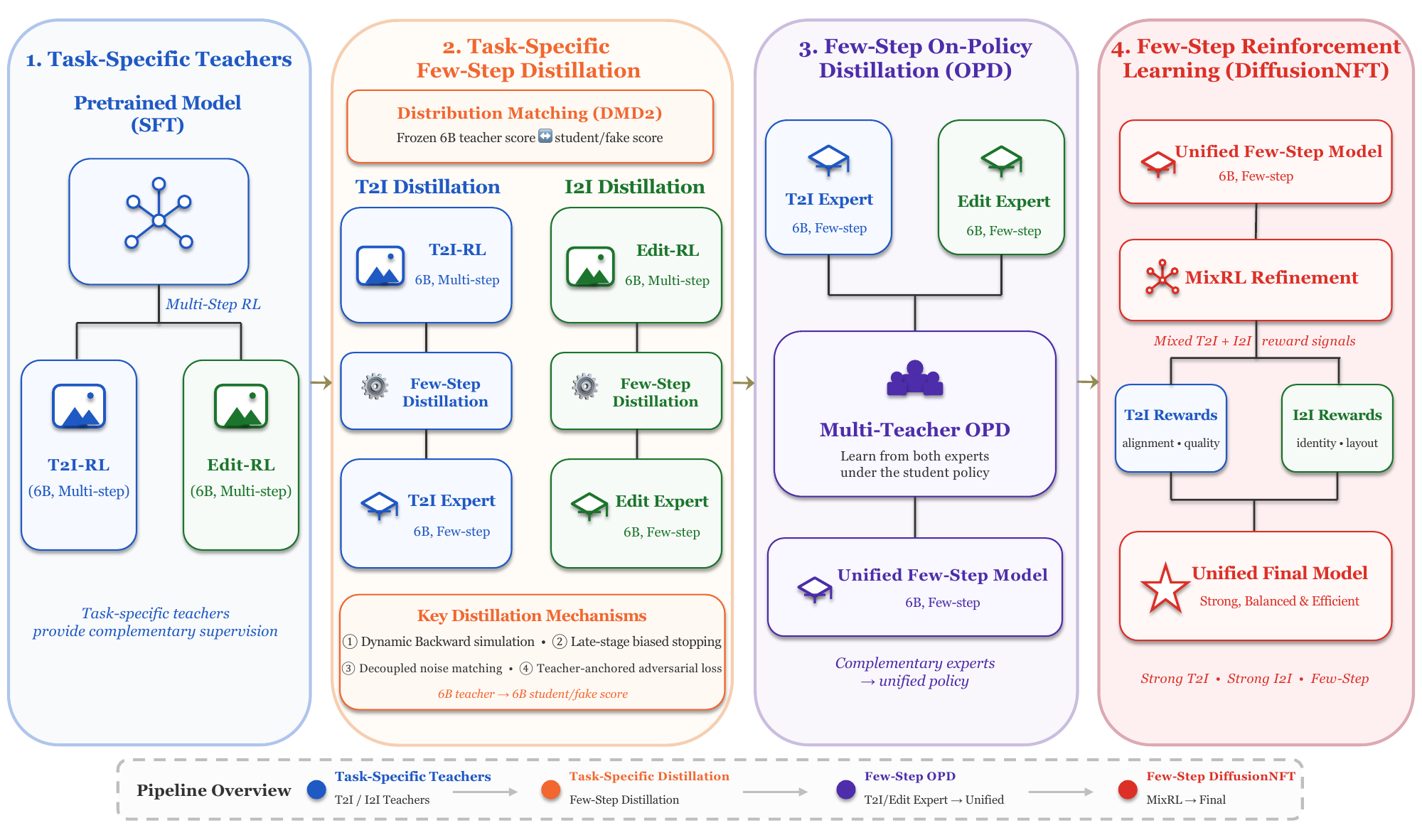}
    \caption{Few-step distillation pipeline. Task-specific RL teachers are compressed using our improved DMD2 recipe, consolidated by multi-teacher OPD, and refined by mixed-task few-step RL.}
    \label{fig:distill}
\end{figure*}
To reduce sampling cost, we distill the multi-step model into a few-step generator, e.g., from 50 to 8 sampling steps, while preserving visual quality, prompt adherence, and structural consistency. Our approach builds on Distribution Matching Distillation (DMD)~\cite{yin2024one,yin2024improved}, which aligns the student's generated distribution with that of a pretrained teacher rather than regressing toward individual teacher samples. Following DMD2~\cite{yin2024improved}, we use a frozen teacher to provide the real score and jointly optimize a fake score network to approximate the student distribution. Their score difference provides the distribution-matching gradient, with multiple fake-score updates performed for each student update to improve optimization stability.

A key challenge in few-step distillation is the mismatch between training states and the student's actual inference trajectory. We mitigate this issue using \textbf{Dynamic Backward Simulation} following CDM~\cite{liu2026continuous}. At each iteration, we randomly sample a step budget from $\{2,4,8,16,28\}$ and partially unroll the student from pure noise to obtain an intermediate state for supervision. Compared with a fixed-step trajectory, randomized budgets expose the student to states produced under different generation qualities and substantially broaden the effective supervision along its own sampling trajectories.

We further introduce three modifications to improve distribution matching. First, we bias the stopping point of backward simulation toward later, lower-noise stages while maintaining coverage of the full trajectory, allocating more supervision to the refinement of textures and fine-grained details. Second, following Decoupled-DMD~\cite{liu2025decoupled}, the generated $x_0$ is re-noised only to timesteps below the stopping noise level. This prevents the teacher from supervising high-noise states that the student has already traversed and concentrates the matching objective on the remaining refinement trajectory. Third, we attach a lightweight adversarial head to a late feature layer of the \emph{frozen teacher}, rather than to the continuously updated fake score as in DMD2. The fixed teacher provides a more stable semantic feature space for discriminating real and student-generated samples, resulting in a more stable adversarial signal.

The formulation also supports heterogeneous teacher--student architectures. When distilling the structurally compressed 3B model in Sec.~\ref{sec:pruning} from the 6B teacher, the fake score adopts the 3B architecture because it models the student's output distribution and is initialized from the student checkpoint. In contrast, the adversarial discriminator remains attached to the frozen 6B teacher, allowing real and generated samples to be compared in the same higher-capacity representation space.

We observe that a single teacher does not provide an optimal few-step student for all tasks: T2I-oriented teachers favor generation quality, whereas editing-oriented teachers better preserve reference consistency and editing fidelity. We therefore distill task-specific few-step experts and subsequently consolidate them using the multi-teacher OPD formulation described above, following the general spirit of DanceOPD~\cite{zhou2026danceopd} and DiffusionOPD~\cite{li2026diffusionopd}. Finally, we apply few-step reinforcement learning with DiffusionNFT~\cite{zheng2025diffusionnft} using mixed T2I and I2I reward signals. This final refinement jointly optimizes text-image alignment, perceptual quality, identity preservation, and structural consistency while retaining the efficiency of few-step sampling.

\paragraph{Recovery after structural pruning.}
The 3B model directly inherits the surviving parameters after reducing the 6B model from 32 to 24 attention heads. It then repeats the coarse-to-fine recovery curriculum through CT instead of receiving only a short recovery run. At SFT, the diffusion objective is supplemented with the velocity-prediction distillation loss in Eq.~\ref{eq:compact_kd}. Finally, task-routed OPD transfers the specialized 6B RL policies into the unified 3B student.

\subsection{Prompt Enhancer Details}
\label{app:prompt_enhancer}

\begin{figure*}[t]
    \centering
    \includegraphics[width=1.0\textwidth]{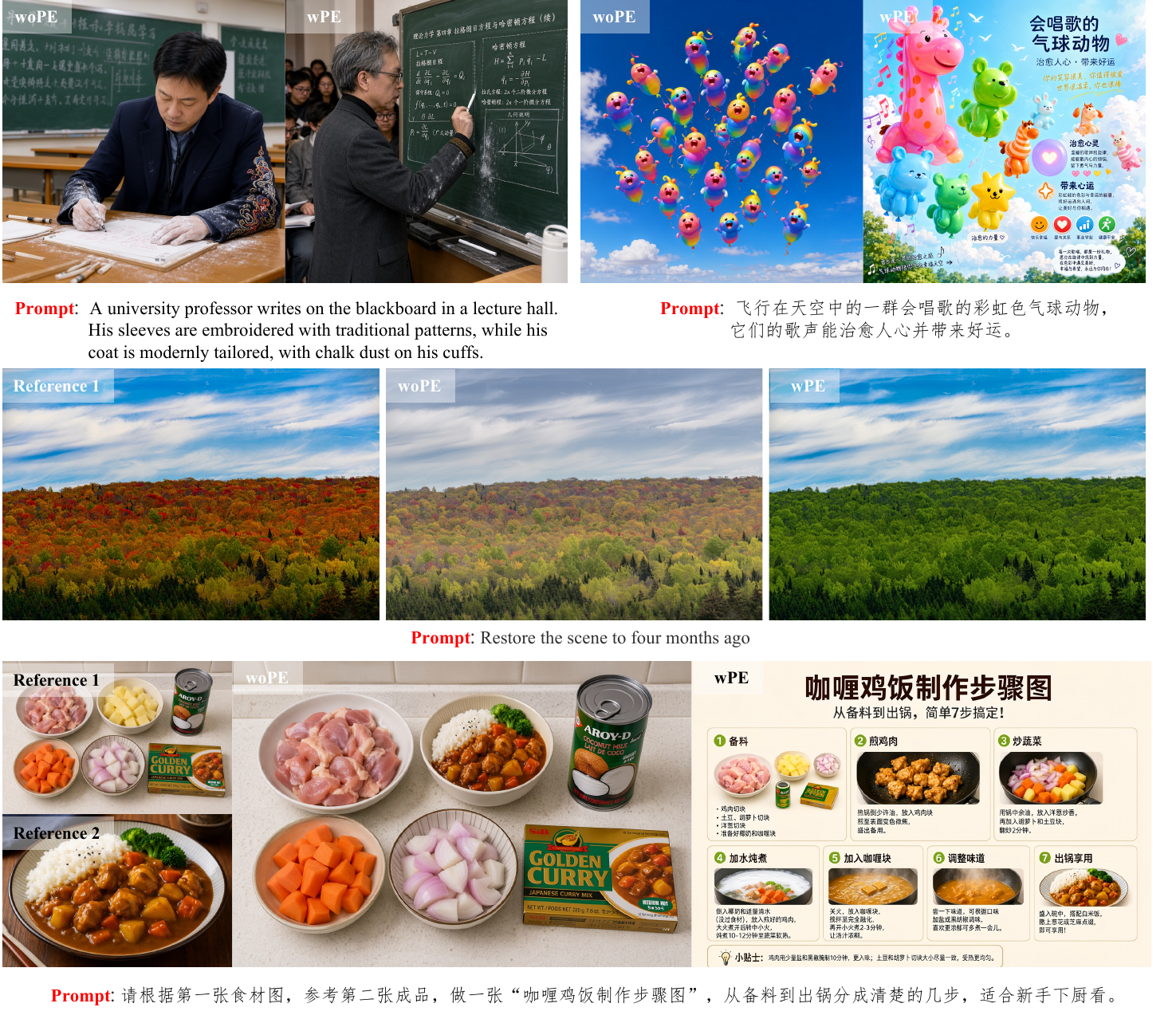}
    \caption{Generation results with and without the Prompt Enhancer on compositionally complex and reasoning-intensive prompts.}
    \label{fig:pe_case}
\end{figure*}
The Prompt Enhancer (PE) converts diverse user instructions into structured, fine-grained prompts that are better aligned with the visual generator, particularly for compositionally complex or knowledge-intensive tasks (Figure~\ref{fig:pe_case}). Rather than treating PE as a conventional prompt rewriter that merely expands short inputs, we formulate it as a learned interface between user intent and visual synthesis.

Conceptually, PE performs four closely related functions.
First, an \textbf{Interpreter} recovers the intended task, target content, reference roles, and modification boundaries from potentially incomplete or ambiguous user inputs.
Second, a \textbf{Reasoner} supplies only the knowledge and implicit constraints necessary to make the request visually realizable.
Third, a \textbf{Planner} converts the resolved intent into an explicit visual plan, including subjects, attributes, spatial relationships, viewpoints, layout regions, text blocks, and preservation requirements.
Finally, a \textbf{Compiler} maps this plan into the vocabulary, descriptive structure, granularity, image-reference convention, and constraint format established by our Captioner.
This final compilation step is essential. During generator training, the Captioner defines the conditional language distribution observed by the DiT. At inference time, the PE Compiler reproduces this language distribution from user requests rather than from target images. Aligning the two distributions reduces the discrepancy between training-time supervision and inference-time user conditions:
\begin{equation}
P_{\mathrm{PE}}(p \mid u,R)
\approx
P_{\mathrm{Captioner}}
(p \mid I_{\mathrm{target}},R).
\label{eq:captioner_pe_alignment}
\end{equation}
where $u$ denotes the user request,
$R=\{I_1,\ldots,I_n\}$ denotes the optional reference-image set,
and $p$ denotes a generator-aligned visual condition.

A single PE supports both text-to-image (T2I) generation and image editing (I2I), and adaptively determines whether explicit reasoning is required: simple requests are rewritten directly, whereas complex requests invoke additional reasoning before producing the final \texttt{refined\_prompt}. For text-to-image generation, the compiled prompt describes the complete target image at a task-appropriate granularity.
For image editing, visual content already specified by the inputs is represented through image references, while newly introduced or modified content is expressed using the same vocabulary and descriptive organization as text-to-image captions. PE training consists of three stages:
data construction, supervised fine-tuning, and generator-aware reinforcement learning.

\subsubsection{Data Construction}

We construct two complementary types of training data: \textbf{Universal} data for broad prompt rewriting and \textbf{Reasoning} data for tasks requiring non-trivial intent understanding, domain knowledge, or spatial planning.

\paragraph{Universal Data.}
We reuse the T2I and I2I training corpus described previously, together with
the \texttt{refined\_prompt} produced by the captioning pipeline for each
training sample. These captioner-generated descriptions define the target
language, granularity, and organization expected by the visual backbone.
Gemini 3.1 Pro subsequently verifies their semantic fidelity and visual
completeness, correcting the \texttt{refined\_prompt} only when necessary and
supplementing it with a reasoning trace for samples that require intent
disambiguation, knowledge reasoning, or spatial planning.

Starting from the verified \texttt{refined\_prompt}, we synthesize diverse
user inputs with different lengths, linguistic styles, levels of detail, and
representations, including natural-language requests, structured
instructions, and programmatic formats. For I2I samples, the synthesis also
incorporates the available reference images and preserves the correspondence
between image-grounded entities and their \texttt{[image N]} references.
The synthesized inputs are further mixed with real user queries. This
construction maps heterogeneous T2I and I2I requests to a consistent,
captioner-aligned visual condition representation, while providing explicit
reasoning supervision only when required by the task.

\paragraph{Reasoning Data.}
We additionally synthesize tasks whose desired visual outcome cannot be obtained through direct surface-level rewriting. For T2I, Claude generates concise reasoning-oriented instructions and their underlying rationale, GPT-Image renders the corresponding target image, and Gemini produces a target \texttt{refined\_prompt} conditioned on the realized image. For I2I, the same procedure is extended with reference images and editing instructions. The resulting data explicitly cover knowledge-intensive and layout-sensitive generation and editing scenarios.

\paragraph{Adaptive Thinking Annotation.}
After construction, we balance Universal/Reasoning and T2I/I2I data across task categories. Each sample is further assigned a reasoning mode according to task complexity. Complex samples retain both the \texttt{<think>} trace and \texttt{refined\_prompt} as supervision, whereas simple samples supervise only the final \texttt{refined\_prompt}. This enables adaptive reasoning without imposing unnecessary inference overhead on straightforward requests.

\subsubsection{Supervised Fine-Tuning}

SFT teaches the PE a unified rewriting space for both generation and editing while initializing its adaptive reasoning behavior. Regardless of the input style or granularity, the model produces a standardized \texttt{refined\_prompt}. For T2I, the output provides a complete fine-grained visual description. For I2I, unchanged content is referenced through the input image, while only requested modifications, newly introduced content, and target visual attributes are explicitly described.

The model simultaneously learns when reasoning is necessary: visually explicit requests are rewritten directly, while knowledge-intensive, ambiguous, or layout-sensitive tasks first produce a \texttt{<think>} trace before the final prompt. This yields a single model that supports heterogeneous T2I/I2I inputs with adaptive computation.
\subsubsection{Generator-Aware Reinforcement Learning}
Starting from the SFT checkpoint, we further optimize the PE with GRPO while keeping the visual DiT frozen. For each user input, the PE samples a group of candidate rewrites with adaptive reasoning. Each final \texttt{refined\_prompt} is evaluated from two complementary perspectives: its textual quality is assessed directly by a \textbf{Rewrite Text Reward}, while the prompt is also rendered by the frozen DiT and evaluated by an \textbf{Image Generation Reward}. Only the PE is updated.

\paragraph{Rewrite Text Reward.}
The text-level reward evaluates whether the rewritten prompt faithfully and explicitly represents the user's intent. It contains two dimensions: \textbf{instruction following (IF)}, covering visual constraints, spatial relations, text rendering requirements, and description completeness; and \textbf{knowledge reasoning (KR)}, evaluating entity recognition, factual visual knowledge, and explicit layout planning for reasoning-intensive tasks.

\paragraph{Image Generation Reward.}
The image-level reward directly evaluates whether the rewritten prompt leads to the desired visual outcome. Its \textbf{instruction-following} component measures realization of the requested content, knowledge, and spatial relationships, while its \textbf{visual-quality (VQ)} component evaluates perceptual quality, aesthetics, and, for editing tasks, consistency with the reference image.

The two views provide complementary supervision: text-level rewards encourage correct and controllable prompt formulation, whereas image-level rewards capture the actual executability of the prompt by the frozen generator. For reasoning-intensive categories, task-specific knowledge and layout criteria are incorporated into both views.

To avoid manually tuning reward scales, we standardize each reward within its GRPO sampling group and fuse the resulting relative advantages. This jointly optimizes semantic correctness, prompt controllability, and final generation quality, while allowing the PE to adapt its rewriting strategy to the capabilities and preferences of the underlying visual generator.

\subsection{Detailed Evaluation Results}
\label{app:detailed_results}
\paragraph{Public editing benchmarks.}
Tables~\ref{tab:gedit_bench}--\ref{tab:rededit} report the
category-wise results on GEdit-Bench, ImgEdit-Bench, and
REDEdit-Bench. Swift-Image shows consistently strong performance
across diverse editing operations, while PE generally provides
further improvements across both the 3B and 6B variants.
The gains are particularly evident on more compositional and
instruction-intensive categories.

\begin{table}[h]
    \centering
    \caption{
        Results on GEdit-Bench.
    }
    \label{tab:gedit_bench}

    \begin{threeparttable}
    \small
    \renewcommand{\arraystretch}{1.2}
        \resizebox{\columnwidth}{!}{
    \begin{tabular}{l| c c c | c c c}
        \toprule
        \multirow{2}{*}{\textbf{Model}} 
        & \multicolumn{3}{c|}{\textbf{GEdit-Bench-EN}} 
        & \multicolumn{3}{c}{\textbf{GEdit-Bench-CN}} \\
        
        \cmidrule(lr){2-4} \cmidrule(lr){5-7} 
        
        & \textbf{G\_SC$\uparrow$} 
        & \textbf{G\_PQ$\uparrow$} 
        & \textbf{G\_O$\uparrow$}
        & \textbf{G\_SC$\uparrow$} 
        & \textbf{G\_PQ$\uparrow$} 
        & \textbf{G\_O$\uparrow$} \\
        \midrule
        
        \rowcolor{groupgray}[0pt][0pt]
        \multicolumn{7}{@{}l@{}}{
            \hspace{0.5em}\textit{\textbf{Proprietary / Closed-source Models}}
        } \\

        GPT-Image-2\cite{gptimage2_model_card}
        & 9.314 & 8.354 & 8.693 & 9.403 & 8.403 & 8.796
 \\

        Nano Banana Pro\cite{google2025nanobanana}
        & 8.102 & 8.344 & 7.738 & 8.135 & 8.306 & 7.799  \\

        Seedream5 Pro\cite{seedream2025seedream}
        & 9.176 & 8.422 & 8.637 & 9.203 & 8.391 & 8.632 \\
        
        Seedream4.5\cite{seedream2025seedream}
        & 8.268 & 8.167 & 7.820 & 8.254 & 8.167 & 7.800 \\

        Qwen Image 2.0 Pro\cite{zhao2026qwen}
        & 9.138 & 8.208 & 8.529 & 9.266 & 8.258 & 8.667 \\
        \midrule
        
        \rowcolor{groupgray}[0pt][0pt]
        \multicolumn{7}{@{}l@{}}{
            \hspace{0.5em}\textit{\textbf{Open-source Models}}
        } \\

        Qwen-Image-Edit-2511~\cite{wu2025qwen}
        & 8.297 & 8.202 & 7.877 & 8.252 & 8.134 & 7.819 \\

        LongCat-Image-Edit~\cite{team2025longcat}
        & 8.128 & 8.177 & 7.748 & 8.141 & 8.117 & 7.731 \\
        
        FLUX.2-klein-9B~\cite{blackforest2025flux2klein}
        & 8.706 & 8.036 & 8.127 & 8.698 & 8.032 & 8.144 \\
        
        FLUX.2-klein-4B~\cite{blackforest2025flux2klein}
        & 8.295 & 8.007 & 7.790 & 8.152 & 7.952 & 7.657 \\
        
        FireRed-Image-Edit~\cite{firered2026rededit}
        & 8.363 & 8.245 & 7.943 & 8.287 & 8.227 & 7.887 \\

        JoyAI-Image-Edit~\cite{song2026joyai}
        & 8.829 & 8.120 & 8.276 & 8.618 & 8.119 & 8.125 \\

        JoyAI-Image-Edit-Plus~\cite{song2026joyai}
        & 8.046 & 7.987 & 7.549 & 7.149 & 8.030 & 6.829 \\

        Boogu-Image-0.1-Edit~\cite{chen2026booguimage01} 
        & 8.700 & 7.712 & 7.905 & 8.525 & 7.771 & 7.763
         \\
        Boogu-Image-0.1-Edit-Thinking~\cite{chen2026booguimage01}
        & 9.042 & 7.900 & 8.246 & 9.000 & 7.847 & 8.229
         \\
        \ours-3B
        & 8.831
        & 7.867
        & 8.106
        & 8.626
        & 7.929
        & 7.972 \\
        
        \ours-3B w/ PE
        & 8.894
        & 7.970
        & 8.174
        & 8.951
        & 8.079
        & 8.317 \\

        \ours-3B w/ PE (API)
        & 8.991
        & 8.100
        & 8.342
        & 9.124
        & 8.100
        & 8.434 \\
        
        \rowcolor{oursred}
        \textbf{\ours-6B}
        & 8.711
        & 7.920
        & 8.028
        & 8.592
        & 7.915
        & 7.940 \\
        
        \rowcolor{oursred}
        \textbf{\ours-6B w/ PE}
        & 8.895
        & 8.190
        & 8.321
        & 8.881
        & 8.228
        & 8.379 \\

        \rowcolor{oursred}
        \textbf{\ours-6B w/ PE (API)}
        & 8.918
        &  8.215
        &  8.350
        &  9.113
        &  8.218
        &  8.493 \\
        
        \bottomrule
    \end{tabular}}
    \end{threeparttable}
\end{table}

\begin{table}[!t]
    \centering
    \caption{
        Results on ImgEdit-Bench.
    }
    \label{tab:imgedit}

    \begin{threeparttable}
    \small
    \setlength{\tabcolsep}{2pt}
    \renewcommand{\arraystretch}{1.2}
        \resizebox{\columnwidth}{!}{
\begin{tabular}{l| c c c c c c c c c| c}
        \toprule
        \textbf{Model}
        & \textbf{Add}
        & \textbf{Adjust}
        & \textbf{Extract}
        & \textbf{Replace}
        & \textbf{Remove}
        & \textbf{Background}
        & \textbf{Style}
        & \textbf{Hybird}
        & \textbf{Action}
        & \textbf{Overall$\uparrow$}\\
        \midrule
        \rowcolor{groupgray}[0pt][0pt]
        \multicolumn{11}{@{}l@{}}{
            \hspace{0.5em}\textit{\textbf{Proprietary / Closed-source Models}}
        } \\

        GPT-Image-2\cite{gptimage2_model_card}
        & 4.82 & 4.87 & 4.31 & 4.93 & 4.74 & 4.86 & 4.99 & 4.43 & 4.78 & 4.74 \\

        Nano Banana Pro\cite{google2025nanobanana}
       & 4.44 & 4.62 & 3.42 & 4.60 & 4.63 & 4.32 & 4.97 & 3.64 & 4.69 & 4.37  \\

        Seedream5 Pro\cite{seedream2025seedream}
        & 4.8 & 4.85 & 3.24 & 4.83 & 4.81 & 4.85 & 4.93 & 3.9 & 4.94 & 4.57  \\
        
        Seedream4.5\cite{seedream2025seedream}
        & 4.57 & 4.65 & 2.97 & 4.66 & 4.46 & 4.37 & 4.92 & 3.71 & 4.56 & 4.32  \\

        Qwen Image 2.0 Pro\cite{zhao2026qwen}
        & 4.69 & 4.73 & 3.32 & 4.7 & 4.76 & 4.72 & 4.84 & 4.03 & 4.3 & 4.45  \\
        \midrule
        \rowcolor{groupgray}[0pt][0pt]
        \multicolumn{11}{@{}l@{}}{
            \hspace{0.5em}\textit{\textbf{Open-source Models}}
        } \\

        Qwen-Image-Edit-2511~\cite{wu2025qwen}
        & 4.54 & 4.57 & 4.13 & 4.70 & 4.46 & 4.36 & 4.89 & 4.16 & 4.81 & 4.51  \\

        LongCat-Image-Edit~\cite{team2025longcat}
        & 4.44 & 4.53 & 3.83 & 4.80 & 4.60 & 4.33 & 4.92 & 3.75 & 4.82 & 4.45  \\
        
        FLUX.2-klein-9B~\cite{blackforest2025flux2klein}
        & 4.68 & 4.81 & 2.02 & 4.67 & 4.73 & 4.59 & 4.9 & 3.71 & 4.78 & 4.32 \\
        
        FLUX.2-klein-4B~\cite{blackforest2025flux2klein}
        & 4.62 & 4.62 & 1.68 & 4.58 & 4.23 & 4.5 & 4.85 & 3.39 & 4.61 & 4.12  \\
        
        FireRed-Image-Edit~\cite{firered2026rededit}
        & 4.55 & 4.66 & 4.34 & 4.75 & 4.58 & 4.5 & 4.85 & 3.39 & 4.61 & 4.12  \\
        
        JoyAI-Image-Edit~\cite{song2026joyai}
        & 4.47 & 4.48 & 4.31 & 4.57 & 4.75 & 4.33 & 4.79 & 3.72 & 4.69 & 4.46  \\
        
        JoyAI-Image-Edit-Plus~\cite{song2026joyai}
        & 4.67 & 4.66 & 1.8 & 4.53 & 4.25 & 4.31 & 4.6 & 3.2 & 4.18 & 4.02  \\

        Boogu-Image-0.1-Edit~\cite{chen2026booguimage01}
        & 4.71 & 4.50 & 3.69 & 4.65 & 4.75 & 4.44 & 4.94 & 4.04 & 4.90 & 4.51  \\

        Boogu-Image-0.1-Edit-Thinking~\cite{chen2026booguimage01}
        & 4.59 & 4.64 & 4.32 & 4.69 & 4.85 & 4.60 & 4.94 & 4.26 & 4.83 & 4.64  \\

        \ours-3B
        & 4.63
        & 4.55
        & 4.54
        & 4.74
        & 4.72
        & 4.62
        & 4.88
        & 3.64
        & 4.76
        & 4.56 \\

        \ours-3B w/ PE
        & 4.68
        & 4.73
        & 4.19
        & 4.67
        & 4.75
        & 4.62
        & 4.78
        & 4.04
        & 4.74
        & 4.58 \\

        \ours-3B w/ PE (API)
        & 4.74
        & 4.74
        & 4.04
        & 4.83
        & 4.91
        & 4.66
        & 4.88
        & 3.82
        & 4.82
        & 4.60 \\

        \rowcolor{oursred}
        \textbf{\ours-6B}
        & 4.56
        & 4.69
        & 4.44
        & 4.63
        & 4.71
        & 4.58
        & 4.93
        & 3.78
        & 4.75
        & 4.56 \\

        \rowcolor{oursred}
        \textbf{\ours-6B w/ PE}
        & 4.68
        & 4.62
        & 4.31
        & 4.71
        & 4.70
        & 4.66
        & 4.85
        & 4.27
        & 4.87
        & 4.63 \\

        \rowcolor{oursred}
        \textbf{\ours-6B w/ PE (API)}
        & 4.76
        & 4.75
        & 4.36
        & 4.84
        & 4.78
        & 4.74
        & 4.87
        & 3.98
        & 4.70
        & 4.64 \\

        \bottomrule
    \end{tabular}}
    \end{threeparttable}
\end{table}

\begin{table}[!t]
    \centering
    \caption{
        Results on REDEdit-Bench-EN.
    }
    \label{tab:rededit}

    \begin{threeparttable}
    \small
    \renewcommand{\arraystretch}{1.2}
    \resizebox{\columnwidth}{!}{
\begin{tabular}{l| c c c c c c c c c c c c c c c| c}
        \toprule
        \textbf{Model}
        & \textbf{Add}
        & \textbf{Adjust}
        & \textbf{BG}
        & \textbf{Beauty}
        & \textbf{Color}
        & \textbf{Compose}
        & \textbf{Extract}
        & \textbf{Potrait}
        & \textbf{Low-level}
        & \textbf{Motion}
        & \textbf{Remove}
        & \textbf{Replace}
        & \textbf{Stylize}
        & \textbf{Text}
        & \textbf{Viewpoint}
        & \textbf{Overall$\uparrow$}\\
        \midrule
        \rowcolor{groupgray}[0pt][0pt]
        \multicolumn{17}{@{}l@{}}{
            \hspace{0.5em}\textit{\textbf{Proprietary / Closed-source Models}}
        } \\

        GPT-Image-2\cite{gptimage2_model_card}
        & 4.84 & 4.65 & 4.83 & 4.61 & 4.7 & 4.54 & 3.93 & 4.93 & 4.83 & 4.97 & 4.54 & 4.84 & 4.99 & 4.79 & 3.69 & 4.65  \\

        Nano Banana Pro\cite{google2025nanobanana}
        & 4.72 & 4.48 & 4.42 & 4.35 & 4.56 & 4.31 & 3.19 & 4.88 & 4.76 & 4.87 & 4.38 & 4.71 & 4.85 & 4.8 & 3.59 & 4.46 \\

        Seedream5 Pro\cite{seedream2025seedream}
        & 4.82 & 4.68 & 4.84 & 4.68 & 4.75 & 4.4 & 3.53 & 4.89 & 4.89 & 4.92 & 4.5 & 4.86 & 4.87 & 4.76 & 3.92 & 4.62  \\
        
        Seedream4.5\cite{seedream2025seedream}
        & 4.56 & 3.95 & 4.65 & 3.94 & 4.07 & 4.11 & 3.59 & 4.88 & 4.73 & 4.86 & 4.21 & 4.57 & 4.88 & 4.38 & 3.3 & 4.31  \\

        Qwen Image 2.0 Pro\cite{zhao2026qwen}
        & 4.63 & 4.39 & 4.62 & 4.28 & 4.42 & 4.16 & 2.75 & 4.74 & 4.89 & 4.72 & 4.47 & 4.73 & 4.7 & 4.46 & 3.67 & 4.38  \\
        \midrule
        \rowcolor{groupgray}[0pt][0pt]
        \multicolumn{17}{@{}l@{}}{
            \hspace{0.5em}\textit{\textbf{Open-source Models}}
        } \\

        Qwen-Image-Edit-2511~\cite{wu2025qwen}
        & 4.55 & 4.17 & 4.56 & 3.49 & 4.07 & 4.07 & 3.54 & 4.42 & 4.52 & 4.72 & 4.42 & 4.65 & 4.85 & 4.06 & 3.38 & 4.23  \\

        LongCat-Image-Edit~\cite{team2025longcat}
        & 4.38 & 4.04 & 4.49 & 3.89 & 4.10 & 3.93 & 2.98 & 4.47 & 4.27 & 4.69 & 4.24 & 4.51 & 4.86 & 3.83 & 3.25 & 4.12  \\
        
        FLUX.2-klein-9B~\cite{blackforest2025flux2klein}
        & 4.52 & 3.86 & 4.59 & 3.71 & 4.08 & 3.76 & 2.52 & 4.45 & 4.54 & 4.53 & 3.92 & 4.54 & 4.87 & 3.48 & 3.66 & 4.07 \\
        
        FLUX.2-klein-4B~\cite{blackforest2025flux2klein}
        & 4.36 & 3.79 & 4.51 & 3.69 & 3.92 & 3.68 & 2.67 & 4.32 & 4.56 & 4.38 & 3.57 & 4.17 & 4.88 & 3.01 & 3.6 & 3.94  \\
        
        FireRed-Image-Edit~\cite{firered2026rededit}
        & 4.41 & 4.33 & 4.60 & 3.55 & 4.47 & 4.25 & 3.49 & 4.50 & 4.44 & 4.65 & 4.46 & 4.70 & 4.94 & 4.44 & 2.78 & 4.26  \\

        JoyAI-Image-Edit~\cite{song2026joyai}
        & 4.43 & 4.01 & 4.57 & 3.11 & 4.13 & 4.01 & 3.61 & 4.31 & 4.64 & 4.25 & 4.33 & 4.61 & 4.92 & 4.34 & 3.24 & 4.17  \\

        JoyAI-Image-Edit-Plus~\cite{song2026joyai}
        & 4.35 & 4.13 & 4.19 & 3.64 & 3.92 & 3.53 & 2.6 & 4.47 & 4.23 & 4.28 & 3.53 & 3.71 & 4.05 & 4.02 & 3.37 & 3.87  \\

        Boogu-Image-0.1-Edit~\cite{chen2026booguimage01}
        & 4.43 & 3.90 & 4.45 & 3.33 & 4.14 & 3.68 & 2.79 & 4.00 & 4.63 & 4.51 & 4.03 & 4.46 & 4.82 & 4.16 & 2.87 & 4.01 \\
        Boogu-Image-0.1-Edit-Thinking~\cite{chen2026booguimage01}
        & 4.38 & 4.32 & 4.42 & 3.95 & 4.08 & 3.84 & 2.86 & 4.42 & 3.62 & 4.72 & 4.27 & 4.61 & 4.92 & 4.32 & 2.95 & 4.11
        \\
        \ours-3B
        & 4.44
        & 4.27
        & 4.61
        & 4.20
        & 4.23
        & 3.85
        & 4.03
        & 4.42
        & 4.57
        & 4.57
        & 4.24
        & 4.47
        & 4.79
        & 4.06
        & 3.57
        & 4.29
        \\
        
        \ours-3B w/ PE
        & 4.55
        & 4.36
        & 4.60
        & 4.47
        & 4.08
        & 4.14
        & 4.01
        & 4.40
        & 4.71
        & 4.69
        & 4.27
        & 4.48
        & 4.70
        & 4.31
        & 3.72
        & 4.37
        \\

        \ours-3B w/ PE (API)
        & 4.55
        & 4.41
        & 4.69
        & 4.52
        & 4.46
        & 4.15
        & 3.99
        & 4.46
        & 4.58
        & 4.64
        & 4.48
        & 4.67
        & 4.83
        & 4.21
        & 3.54
        & 4.41
        \\
        
        \rowcolor{oursred}
        \textbf{\ours-6B}
        & 4.47
        & 4.25
        & 4.63
        & 4.31
        & 4.30
        & 4.01
        & 4.05
        & 4.38
        & 4.68
        & 4.58
        & 4.16
        & 4.56
        & 4.91
        & 4.06
        & 3.71
        & 4.34
        \\
        
        \rowcolor{oursred}
        \textbf{\ours-6B w/ PE}
        & 4.54
        & 4.35
        & 4.61
        & 4.46
        & 4.35
        & 4.02
        & 3.97
        & 4.45
        & 4.74
        & 4.81
        & 4.27
        & 4.60
        & 4.75
        & 4.28
        & 3.59
        & 4.39
        \\

        \rowcolor{oursred}
        \textbf{\ours-6B w/ PE (API)}
        & 4.70
        & 4.48
        & 4.71
        & 4.56
        & 4.37
        & 4.10
        & 4.21
        & 4.49
        & 4.59
        & 4.65
        & 4.38
        & 4.71
        & 4.83
        & 4.22
        & 3.70
        & 4.45
        \\

        \bottomrule
    \end{tabular}}
    \end{threeparttable}
\end{table}

\paragraph{General single- and multi-image editing.}
Tables~\ref{tab:pi_general_single} and
\ref{tab:pi_general_multi} provide the detailed results on
CPI-General. Swift-Image performs strongly across both single-
and multi-image editing, with PE yielding especially pronounced
gains in the multi-image setting, where resolving reference roles
and compositional relations is more challenging.

\begin{table}[!t]
    \centering
    \caption{
        Results on CPI-General-Benchmark-EN(Single-image editing).
    }
    \label{tab:pi_general_single}

    \begin{threeparttable}
    \small
    \renewcommand{\arraystretch}{1.2}
    \resizebox{\columnwidth}{!}{
\begin{tabular}{l| c c c c c c c c c c c c c c| c}
        \toprule
        \textbf{Model}
        & \textbf{Add}
& \textbf{Adjust}
& \textbf{Compose}
& \textbf{Extract}
& \textbf{Low Level}
& \textbf{Motion}
& \textbf{Ps Human}
& \textbf{Remove}
& \textbf{Replace}
& \textbf{Stylize}
& \textbf{Structure-Guided}
& \textbf{Subject-Driven}
& \textbf{Text}
& \textbf{Viewpoint}
        & \textbf{Overall$\uparrow$}\\
        \midrule
        \rowcolor{groupgray}[0pt][0pt]
        \multicolumn{16}{@{}l@{}}{
            \hspace{0.5em}\textit{\textbf{Proprietary / Closed-source Models}}
        } \\

        GPT-Image-2\cite{gptimage2_model_card}
        & 4.56 & 4.69 & 4.69 & 4.54 & 4.72 & 4.84 & 4.64 & 4.83 & 4.77 & 4.78 & 5.00 & 4.83 & 4.53 & 4.65 & 4.74
  \\

        Nano Banana Pro\cite{google2025nanobanana}
        & 4.71 & 4.45 & 4.48 & 4.04 & 4.65 & 4.49 & 4.58 & 4.40 & 4.80 & 4.86 & 4.74 & 4.82 & 4.81 & 4.55 & 4.61
  \\

        Seedream5 Pro\cite{seedream2025seedream}
        & 4.78 & 4.65 & 4.68 & 4.10 & 4.58 & 4.79 & 4.71 & 4.56 & 4.95 & 4.84 & 4.77 & 4.84 & 4.87 & 4.45 & 4.71
  \\
        
        Seedream4.5\cite{seedream2025seedream}
        & 4.54 & 4.36 & 4.43 & 4.26 & 4.36 & 4.57 & 4.46 & 4.28 & 4.68 & 4.84 & 4.66 & 4.75 & 4.49 & 4.17 & 4.49  \\

        Qwen Image 2.0 Pro\cite{zhao2026qwen}
        & 4.70 & 4.41 & 4.47 & 3.25 & 4.44 & 4.59 & 4.57 & 4.43 & 4.80 & 4.89 & 4.37 & 4.69 & 4.71 & 4.58 & 4.55  \\
        \midrule
        \rowcolor{groupgray}[0pt][0pt]
        \multicolumn{16}{@{}l@{}}{
            \hspace{0.5em}\textit{\textbf{Open-source Models}}
        } \\

        Qwen-Image-Edit-2511~\cite{wu2025qwen}
        & 4.39 & 4.14 & 4.15 & 3.63 & 4.09 & 4.54 & 4.10 & 4.37 & 4.47 & 4.79 & 3.83 & 4.38 & 4.39 & 4.31 & 4.30  \\

        
        FLUX.2-klein-9B~\cite{blackforest2025flux2klein}
        & 4.46 & 4.14 & 3.96 & 2.46 & 3.99 & 4.08 & 4.03 & 4.31 & 4.53 & 4.83 & 4.36 & 4.62 & 3.69 & 4.62 & 4.18 \\
        
        FLUX.2-klein-4B~\cite{blackforest2025flux2klein}
        & 4.28 & 4.15 & 4.09 & 3.05 & 3.48 & 3.96 & 3.96 & 4.16 & 4.37 & 4.61 & 3.75 & 4.39 & 3.25 & 4.28 & 4.03  \\
        
        FireRed-Image-Edit~\cite{firered2026rededit}
        & 4.39 & 4.39 & 4.23 & 3.58 & 4.32 & 4.60 & 4.01 & 4.44 & 4.76 & 4.96 & 3.46 & 4.29 & 4.55 & 4.29 & 4.38  \\

        JoyAI-Image-Edit-Plus~\cite{song2026joyai}
        & 4.35 & 4.08 & 3.83 & 2.19 & 2.90 & 4.05 & 3.64 & 3.73 & 4.24 & 3.70 & 2.55 & 3.84 & 4.20 & 3.84 & 3.81  \\

        \ours-3B
        & 4.33
        & 4.39
        & 4.17
        & 4.26
        & 3.81
        & 4.30
        & 4.27
        & 4.24
        & 4.54
        & 4.78
        & 4.09
        & 4.64
        & 4.38
        & 4.37
        & 4.35
        \\
        
        \ours-3B w/ PE
        & 4.47
        & 4.49
        & 4.22
        & 4.33
        & 4.26
        & 4.62
        & 4.54
        & 4.32
        & 4.70
        & 4.83
        & 4.54
        & 4.59
        & 4.60
        & 4.24
        & 4.49
        \\

        \ours-3B w/ PE (API)
        & 4.46
        & 4.55
        & 4.33
        & 4.64
        & 4.37
        & 4.58
        & 4.71
        & 4.49
        & 4.59
        & 4.87
        & 4.94
        & 4.66
        & 4.50
        & 4.66
        & 4.57
        \\
        
        \rowcolor{oursred}
        \textbf{\ours-6B}
        & 4.39
        & 4.34
        & 4.19
        & 4.36
        & 3.82
        & 4.33
        & 4.35
        & 4.13
        & 4.40
        & 4.77
        & 3.22
        & 4.50
        & 4.35
        & 4.09
        & 4.30
        \\
        
        \rowcolor{oursred}
        \textbf{\ours-6B w/ PE}
        & 4.57
        & 4.47
        & 4.23
        & 4.40
        & 4.25
        & 4.45
        & 4.58
        & 4.34
        & 4.61
        & 4.86
        & 4.69
        & 4.64
        & 4.52
        & 4.54
        & 4.50
        \\

        \rowcolor{oursred}
        \textbf{\ours-6B w/ PE (API)}
        & 4.53
        & 4.49
        & 4.34
        & 4.74
        & 4.56
        & 4.52
        & 4.66
        & 4.45
        & 4.67
        & 4.95
        & 5.00
        & 4.70
        & 4.61
        & 4.77
        & 4.60
        \\

        \bottomrule
    \end{tabular}}
    \end{threeparttable}
\end{table}

\begin{table}[!t]
    \centering
    \caption{
        Results on CPI-General-Benchmark-EN(Multi-image editing).
    }
    \label{tab:pi_general_multi}

    \begin{threeparttable}
    \small
    \renewcommand{\arraystretch}{1.2}
    \resizebox{\columnwidth}{!}{
\begin{tabular}{l| c c c c c c c c c c| c}
        \toprule
        \textbf{Model}
        & \textbf{Multi-Image Compose}
& \textbf{Multi-Subject-Driven}
& \textbf{Ref-Add}
& \textbf{Ref-Change}
& \textbf{Ref-Motion}
& \textbf{Ref-Remove}
& \textbf{Ref-Replace}
& \textbf{Ref-Stylize}
& \textbf{Ref-Text}
& \textbf{Ref-Viewpoint}

        & \textbf{Overall$\uparrow$}\\
        \midrule
        \rowcolor{groupgray}[0pt][0pt]
        \multicolumn{12}{@{}l@{}}{
            \hspace{0.5em}\textit{\textbf{Proprietary / Closed-source Models}}
        } \\

        GPT-Image-2\cite{gptimage2_model_card}
        & 4.65 & 4.80 & 4.56 & 4.36 & 4.20 & 4.03 & 4.60 & 4.75 & 4.42 & 4.60
        & 4.51 \\

        Nano Banana Pro\cite{google2025nanobanana}
        & 4.34 & 4.48 & 4.24 & 4.27 & 4.37 & 4.11 & 4.28 & 4.48 & 4.10 & 3.79 & 4.27  \\

        Seedream5 Pro\cite{seedream2025seedream}
        & 4.60 & 4.84 & 4.69 & 4.43 & 4.30 & 4.39 & 4.68 & 4.35 & 4.39 & 3.62 & 4.50  \\
        
        Seedream4.5\cite{seedream2025seedream}
        & 3.92 & 4.43 & 4.30 & 3.94 & 3.71 & 4.21 & 4.24 & 4.10 & 3.97 & 2.89 & 4.06  \\

        Qwen Image 2.0 Pro\cite{zhao2026qwen}
        & 4.18 & 4.56 & 4.17 & 4.19 & 3.91 & 4.03 & 4.23 & 4.25 & 3.82 & 3.46 & 4.10  \\
        \midrule
        \rowcolor{groupgray}[0pt][0pt]
        \multicolumn{12}{@{}l@{}}{
            \hspace{0.5em}\textit{\textbf{Open-source Models}}
        } \\

        Qwen-Image-Edit-2511~\cite{wu2025qwen}
        & 2.75 & 3.62 & 3.91 & 3.33 & 3.01 & 2.95 & 3.65 & 2.10 & 3.11 & 1.96 & 3.13  \\

        
        FLUX.2-klein-9B~\cite{blackforest2025flux2klein}
        & 3.45 & 3.65 & 4.04 & 3.70 & 2.87 & 3.03 & 3.75 & 2.88 & 2.95 & 2.97 & 3.38 \\
        
        FLUX.2-klein-4B~\cite{blackforest2025flux2klein}
        & 3.45 & 3.65 & 4.04 & 3.70 & 2.87 & 3.03 & 3.75 & 2.88 & 2.95 & 2.97 & 3.38  \\
        
        FireRed-Image-Edit~\cite{firered2026rededit}
        & 2.58 & 3.13 & 3.83 & 3.20 & 3.47 & 3.73 & 3.21 & 2.49 & 2.76 & 2.18 & 3.11  \\

        JoyAI-Image-Edit-Plus~\cite{song2026joyai}
        & 2.70 & 3.52 & 3.78 & 3.06 & 2.51 & 2.02 & 3.41 & 1.82 & 2.93 & 2.49 & 2.89  \\

        \ours-3B
        & 3.58
        & 3.86
        & 3.92
        & 3.36
        & 3.37
        & 2.70
        & 4.01
        & 2.68
        & 3.29
        & 2.24
        & 3.39
        \\
        
        \ours-3B w/ PE
        & 3.71
        & 4.08
        & 3.93
        & 4.00
        & 3.82
        & 4.16
        & 4.17
        & 3.73
        & 3.85
        & 3.94
        & 3.95
        \\

        \ours-3B w/ PE (API)
        & 3.81
        & 4.28
        & 4.24
        & 4.33
        & 3.93
        & 3.20
        & 4.24
        & 4.53
        & 4.00
        & 4.01
        & 4.05
        \\
        
        \rowcolor{oursred}
        \textbf{\ours-6B}
        & 3.67
        & 4.23
        & 3.94
        & 3.08
        & 3.28
        & 3.06
        & 4.23
        & 2.51
        & 3.39
        & 2.31
        & 3.49
        \\
        
        \rowcolor{oursred}
        \textbf{\ours-6B w/ PE}
        & 3.97
        & 4.19
        & 4.09
        & 3.68
        & 3.53
        & 4.03
        & 4.18
        & 4.00
        & 3.78
        & 3.72
        & 3.96
        \\

        \rowcolor{oursred}
        \textbf{\ours-6B w/ PE (API)}
        & 4.02
        & 4.40
        & 4.16
        & 4.04
        & 4.20
        & 3.36
        & 4.26
        & 4.44
        & 4.04
        & 4.04
        & 4.09
        \\

        \bottomrule
    \end{tabular}}
    \end{threeparttable}
\end{table}
\paragraph{Practical editing scenarios.}
Table~\ref{tab:CPI-life-benchmark} further breaks down performance
on CPI-Practical. PE substantially strengthens Swift-Image on
application-oriented tasks, and the 3B variant remains highly
competitive with the 6B model despite its smaller backbone.
\begin{table}[!t]
    \centering
    \caption{
        Results on CPI-Practical-benchmark-EN.
    }
    \label{tab:CPI-life-benchmark}

    \begin{threeparttable}
    \small
    \renewcommand{\arraystretch}{1.2}
    \resizebox{\columnwidth}{!}{
\begin{tabular}{l| c c c c c c c c c c c c | c}
        \toprule
         \textbf{Model}
& \textbf{Ad Create}
& \textbf{Appe. Chg}
& \textbf{Basic Edit}
& \textbf{Outfit}
& \textbf{Hard Fin.}
& \textbf{Lighting}
& \textbf{Pose/View}
& \textbf{Retouch}
& \textbf{Render}
& \textbf{Soft Furn.}
& \textbf{Style and Creative Gen.}
& \textbf{Translation}
        & \textbf{Overall$\uparrow$}\\

        \midrule
        \rowcolor{groupgray}[0pt][0pt]
        \multicolumn{14}{@{}l@{}}{
            \hspace{0.5em}\textit{\textbf{Proprietary / Closed-source Models}}
        } \\

        GPT-Image-2\cite{gptimage2_model_card}
        & 4.74 & 4.49 & 4.41 & 4.87 & 4.96 & 4.80 & 4.68 & 4.70 & 5.00 & 5.00 & 4.89 & 4.60 & 4.69  \\

        Nano Banana Pro\cite{google2025nanobanana}
        & 4.89 & 4.59 & 4.36 & 3.94 & 4.53 & 5.00 & 4.18 & 4.61 & 5.00 & 4.56 & 4.77 & 4.26 & 4.58  \\

        Seedream5 Pro\cite{seedream2025seedream}
        & 4.97 & 4.74 & 4.67 & 4.98 & 4.81 & 4.80 & 4.63 & 4.46 & 4.85 & 4.78 & 4.78 & 4.63 & 4.72  \\
        
        Seedream4.5\cite{seedream2025seedream}
        & 4.34 & 4.30 & 4.09 & 4.57 & 3.75 & 4.80 & 3.63 & 4.53 & 4.73 & 4.40 & 4.54 & 3.73 & 4.32  \\

        Qwen Image 2.0 Pro\cite{zhao2026qwen}
        & 4.07 & 4.48 & 4.48 & 4.75 & 4.38 & 4.43 & 4.31 & 4.12 & 4.80 & 4.55 & 4.48 & 2.86 & 4.39  \\
        \midrule
        \rowcolor{groupgray}[0pt][0pt]
        \multicolumn{14}{@{}l@{}}{
            \hspace{0.5em}\textit{\textbf{Open-source Models}}
        } \\

        Qwen-Image-Edit-2511~\cite{wu2025qwen}
        & 2.73 & 3.93 & 4.16 & 3.44 & 3.41 & 3.70 & 4.06 & 3.86 & 4.49 & 3.92 & 4.45 & 1.23 & 3.85  \\

        
        FLUX.2-klein-9B~\cite{blackforest2025flux2klein}
        & 2.37 & 3.92 & 4.29 & 4.01 & 4.04 & 2.80 & 4.33 & 3.40 & 4.96 & 4.01 & 4.27 & 1.13 & 3.85 \\
        
        FLUX.2-klein-4B~\cite{blackforest2025flux2klein}
        & 2.09 & 3.69 & 4.04 & 3.88 & 3.79 & 2.10 & 3.98 & 3.44 & 4.60 & 3.70 & 3.95 & 1.46 & 3.64  \\
        
        FireRed-Image-Edit~\cite{firered2026rededit}
        & 2.96 & 4.04 & 4.41 & 3.20 & 3.61 & 3.70 & 3.88 & 3.95 & 4.88 & 3.09 & 3.94 & 1.33 & 3.89  \\

        JoyAI-Image-Edit-Plus~\cite{song2026joyai}
        & 2.75 & 3.81 & 3.68 & 3.78 & 3.51 & 2.33 & 3.26 & 3.72 & 4.08 & 4.11 & 2.72 & 1.03 & 3.52  \\

        \ours-3B
        & 2.74
        & 4.20
        & 4.38
        & 4.61
        & 3.68
        & 3.00
        & 3.92
        & 4.22
        & 4.35
        & 3.53
        & 3.32
        & 1.40
        & 3.91
        \\
        
        \ours-3B w/ PE
        & 3.48
        & 4.46
        & 4.36
        & 4.50
        & 3.68
        & 4.90
        & 4.47
        & 4.06
        & 4.78
        & 4.23
        & 3.99
        & 3.77
        & 4.23
        \\

        \ours-3B w/ PE (API)
        & 4.59
        & 4.58
        & 4.46
        & 4.89
        & 4.23
        & 4.83
        & 4.37
        & 4.42
        & 4.75
        & 4.46
        & 4.27
        & 3.33
        & 4.47
        \\
        
        \rowcolor{oursred}
        \textbf{\ours-6B}
        & 3.08
        & 4.13
        & 4.25
        & 4.54
        & 3.57
        & 2.30
        & 4.32
        & 4.23
        & 4.30
        & 3.51
        & 3.62
        & 1.10
        & 3.92
        \\
        
        \rowcolor{oursred}
        \textbf{\ours-6B w/ PE}
        & 3.40
        & 4.34
        & 4.52
        & 4.43
        & 4.18
        & 4.70
        & 4.42
        & 3.81
        & 4.90
        & 4.03
        & 4.10
        & 3.93
        & 4.19
        \\

        \rowcolor{oursred}
        \textbf{\ours-6B w/ PE (API)}
        & 4.74
        & 4.49
        & 4.31
        & 4.94
        & 4.47
        & 4.70
        & 4.43
        & 4.35
        & 4.85
        & 4.50
        & 4.25
        & 3.47
        & 4.44
        \\

        \bottomrule
    \end{tabular}}
    \end{threeparttable}
\end{table}

\end{document}